\documentclass[lettersize,journal,twoside]{IEEEtran}
\usepackage{amsmath,amsfonts}
\usepackage{algorithmic}
\usepackage{algorithm}
\usepackage{array}
\usepackage[caption=false,font=normalsize,labelfont=sf,textfont=sf]{subfig}
\usepackage{textcomp}
\usepackage{stfloats}
\usepackage{url}
\usepackage{verbatim}
\usepackage{graphicx}
\usepackage{cite}
\usepackage{caption}
\usepackage{subcaption}
\usepackage{graphicx}
\usepackage{multirow}
\usepackage{bm}
\usepackage{float}
\floatstyle{plaintop}
\restylefloat{table}
\usepackage{enumitem}

\usepackage{amsmath}
\usepackage{amsfonts}
\usepackage{tabularx}
\usepackage{array}
\usepackage{color}
\usepackage{enumitem}
\usepackage{makecell}
\usepackage{xcolor}
\usepackage{colortbl}
\usepackage{booktabs}

\definecolor{my_red}{RGB}{254, 227, 228}
\definecolor{my_green}{RGB}{227, 254, 223}
\definecolor{my_blue}{RGB}{237, 238, 254}

\newcolumntype{C}{>{\centering\arraybackslash}X}
\newcolumntype{B}{>{\columncolor{my_blue}}C}

\newcommand{\system}{IMG2IMU}

\newcommand{\revised}[1]{{\leavevmode{#1}}}

\newcommand{\eg}{\textit{e.g.},}
\newcommand{\ie}{\textit{i.e.},}
\newcommand{\cf}{\textit{c.f.},}

\begin{document}
\title{\system{}: Translating Knowledge from Large-Scale Images to IMU Sensing Applications}

\author{Hyungjun Yoon, Hyeongheon Cha, Hoang C. Nguyen, Taesik Gong, and Sung-Ju Lee 
\thanks{Hyungjun Yoon, Hyeongheon Cha, and Sung-Ju Lee are with the School of Electrical Engineering, KAIST, Republic of Korea. (e-mail: hyungjun.yoon@kaist.ac.kr; hyeongheon@kaist.ac.kr; profsj@kaist.ac.kr)}
\thanks{Hoang C. Nguyen is with VinUni-Illinois Smart Health Center, Hanoi, Vietnam. (e-mail: hoang.nc@vinuni.edu.vn)}
\thanks{Taesik Gong is with Nokia Bell Labs, Cambridge, UK. (e-mail: taesik.gong@nokia-bell-labs.com)}
}



\maketitle

\begin{abstract}
Pre-training representations acquired via self-supervised learning could achieve high accuracy on even tasks with small training data. Unlike in vision and natural language processing domains, pre-training for IMU-based applications is challenging, as there are few public datasets with sufficient size and diversity to learn generalizable representations. To overcome this problem, we propose \system{} that adapts pre-trained representation from large-scale images to diverse IMU sensing tasks. We convert the sensor data into visually interpretable spectrograms for the model to utilize the knowledge gained from vision. \revised{We further present a sensor-aware pre-training method for images that enables models to acquire particularly impactful knowledge for IMU sensing applications. This involves using contrastive learning on our augmentation set customized for the properties of sensor data.
Our evaluation with four different IMU sensing tasks shows that \system{} outperforms the baselines pre-trained on sensor data by an average of 9.6\%p F1-score,} illustrating that vision knowledge can be usefully incorporated into IMU sensing applications where only limited training data is available.
\end{abstract}

\begin{IEEEkeywords}
Mobile sensing, deep learning, self-supervised learning, contrastive learning
\end{IEEEkeywords}

\section{Introduction}

\IEEEPARstart{N}{umerous} ubiquitous applications utilize deep learning with mobile data collected in everyday life. Motion sensing with inertial measurement units~(IMU), such as accelerometers, has emerged as a significant strand of mobile sensing thanks to its vast array of applications
\revised{, such as activity recognition~\cite{kwapisz2011activity, chavarriaga2013opportunity}, transportations~\cite{carlos2019smartphone, gonzalez2017learning}, agricultures~\cite{kamminga2018robust}, mechanics~\cite{ismail2019review}, and healthcare~\cite{kwon2011validation}.} Sensing applications typically employ deep learning models trained through supervision from task-specific datasets. For such settings, the model's performance depends heavily on the quantity of training data. However, acquiring a large amount of data in IMU sensing is challenging due to the data collection cost, device/user heterogeneity, and privacy concerns. 

Recent research has delved into effectively training deep learning models with limited training data, such as pre-training models to teach knowledge for general tasks~(i.e., representation learning), then fine-tuning them with data from downstream tasks~\cite{bengio2013representation}. A promising method for pre-training the model is self-supervised learning, which uses large amounts of unlabeled data to learn the data characteristics with the predefined pretext task. This strategy performed remarkably well in domains with large public datasets. \revised{As an example in natural language processing, models pre-trained without labels from the abundant Internet texts, such as BERT~\cite{devlin2018bert}, PaLM~\cite{chowdhery2023palm}, and GPT series~\cite{brown2020language}, are used as foundation models for various tasks. In computer vision, models derived from pre-training on large-scale datasets (\eg{} ImageNet~\cite{russakovsky2015imagenet}, COCO~\cite{lin2014microsoft}, and LAION-5B~\cite{schuhmann2022laion}) have achieved state-of-the-art performance on several tasks~\cite{chen2020big}.}

Applying self-supervised learning to IMU sensing showed that pre-training with unlabeled IMU sensor data improves downstream performance~\cite{haresamudram2022assessing}. Prior works, however, focus mainly on Human Activity Recognition~(HAR) tasks. 
The main reason for this concentration could be the lack of variety and quantity from publicly available IMU sensor data. \revised{Contrary to images and texts where massive-scale public datasets exist, 
publicly available large-scale sensing datasets~\cite{chan2021capture, gershuny2020testing, willetts2018statistical, doherty2017large} are centered on HAR with limited diversity. For instance, Capture-24~\cite{chan2021capture} is collected only from the user’s smartwatch at a single sampling rate and hence lacks diversity in the sensing device's type, position, and signal processing method. 
A pre-trained model trained on limited dataset results in a generalizability issue, making it difficult to adapt to downstream tasks with different target tasks, subjects, sensors, and data collection methods (\eg{} sensor position and sampling frequency).}

Motivated by this challenge, we ask: \emph{should the representation for sensors be learned only from the sensor data?} The data collected by sensors could be represented in the form of images, such as spectrograms~\cite{jiang2015human, alsheikh2016deep, hur2018iss2image}. When the data is transformed into 2D, the interpretation of informative features in visual form can be supported by the knowledge of interpreting images as pattern and color recognitions. Considering that generalizable knowledge can be obtained by pre-trained vision models on large-scale image datasets, we propose \system{} that translates the knowledge from vision models to IMU sensing tasks using 2D-transformed sensor data as input.

\IEEEpubidadjcol

In \system{}, sensor data is represented as spectrograms, a popularly used 2D transformation. Three sensor axes are mapped to the RGB channels of the spectrogram image. Our channel mapping was inspired by the previous research that successfully transformed sensor data into a 2D format~\cite{alsheikh2016deep, hur2018iss2image}. In subsequent steps, the pre-trained vision model from large-scale images is used to fine-tune the sensing task with the converted sensor data. Note that a domain gap exists between the knowledge required to interpret the sensor data and the images derived from public datasets. For example, rotating an image 90 degrees to the left depicts the same image from a different viewpoint in vision; however, the frequency and time axes are reversed in the spectrogram image. 

To minimize such a gap, 
we design a pre-training method to learn a representation appropriate for IMU sensing. 
We utilize contrastive learning using specially designed sensor-aware augmentations. We suggest four image augmentations that generate sensor-aware positive samples: \texttt{TranslateX}, \texttt{PermuteX}, \texttt{Hue}, and \texttt{Jitter}. These augmentations teach the model the sensory properties during pre-training. 

With sensor-aware augmentations, we evaluated \system{} on various IMU sensing tasks and found it consistently outperforms the baselines when there was limited training data. When evaluated on a diverse range of IMU sensing tasks, \system{} showed \revised{$9.6\%$p} higher performance in mean F1-score compared with existing self-supervised learning methods designed for HAR.

The main contributions of the paper are as follows:
\begin{itemize}
\item We propose \system{} that utilizes a model pre-trained on a large-scale image dataset and translates it into IMU sensing applications through self-supervised learning on limited sensor data.
\item Based on the domain knowledge of sensors, we present image augmentations that enable contrastive learning on images to learn valuable knowledge for sensing tasks.
\item We analyze how each augmentation affects the model to be robust against sensory properties.
\item We demonstrate through experiments that \system{} improves the performance of diverse sensing tasks where only limited data is available for training.
\end{itemize}

\section{Background and Motivation}

\begin{figure}
    \centering
    \includegraphics[width=0.95\columnwidth{}]{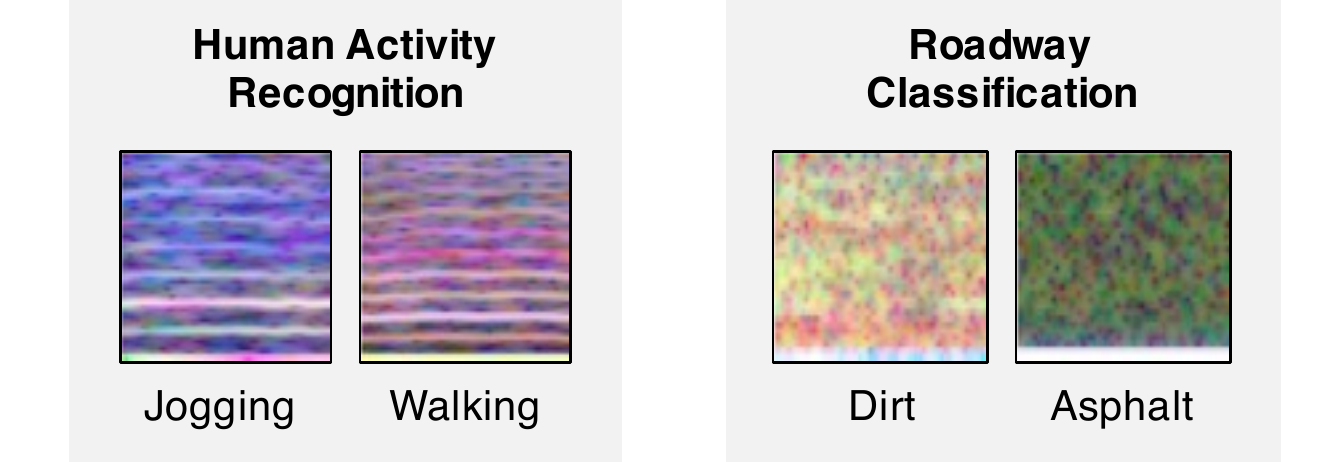}
    \setlength{\belowcaptionskip}{-8pt}
    \caption{Spectrogram images converted from sensor data of human activity recognition and roadway classification tasks.}
    \label{fig:spectrograms}
\end{figure}

\subsection{\revised{From Scarce Sensor Data to Abundant Image Data}}

\revised{
Popular IMU sensing datasets are centered on human motions, such as activities~\cite{kwapisz2011activity, chavarriaga2013opportunity}, gait~\cite{bachlin2009wearable}, and fall detection~\cite{casilari2017umafall}. Although datasets for specific tasks or subjects, \eg{} workout activities~\cite{stromback2020mm}, sports~\cite{brunner2019swimming}, and cattle motion~\cite{kamminga2018robust}, are accessible, they typically lack scale. 
While large-scale efforts such as Capture-24~\cite{chan2021capture} and UK-Biobank~\cite{doherty2017large} exist, their scope is limited to daily human activities measured via wrist-worn devices, thus without diversity in tasks, sensor locations, and subjects. In our evaluation (Section~\ref{section:end-to-end}), we demonstrate that models pre-trained on such dataset (Capture-24) exhibit suboptimal performance when applied to data with different sensor positions~\cite{kwapisz2011activity, bachlin2009wearable}, tasks and subjects~\cite{kamminga2018robust, menegazzo2020multi}. 
We emphasize the shortfall in the sensory domain: the lack of publicly available datasets that provide a comprehensive and diverse collection. The applications of IMU-based sensing are boundless, encompassing healthcare, sports, the automotive industry, and beyond. However, the deficiency in the public datasets limits the models' generalization across various tasks and domains.

On the other hand, computer vision has dramatically benefited from the availability of abundant public data. Improving model performance by enriching the data scale and providing a solid source of pre-trained knowledge has become ubiquitous~\cite{dosovitskiy2020image, kolesnikov2020big, zhai2022scaling}. Beginning with ImageNet~\cite{russakovsky2015imagenet}, which contains 1.2~M images with 1K classes, million-level datasets such as CC12M~\cite{changpinyo2021conceptual} and YFCC100M~\cite{thomee2016yfcc100m} have been introduced. Subsequently, industrial researchers have released even larger datasets, such as Instagram-1B~\cite{yalniz2019billion} 
and JFT3B~\cite{zhai2022scaling}, with a billion or more images. A recent publication on this trend is LAION-5B~\cite{schuhmann2022laion}, which comprised 5.85B images and was shown to produce powerful results. 

These image data in the public domain offer a valuable source of pre-trained knowledge for various applications. For instance, ImageNet pre-trained models have proven effective in dermatology and chest X-ray classification~\cite{azizi2021big}, and even in sound classification~\cite{shin2021self} by transforming audio into mel-spectrograms. These examples highlight the models' exceptional capacity to generalize. 
We investigate the potential of utilizing abundant image data from the public domain to interpret visualized IMU sensor data. We compare this strategy against models pre-trained on large sensor datasets, 
investigating the viability of models pre-trained on image data as a complementary approach for IMU sensing applications. 

}

\begin{figure}
    \centering
    \includegraphics[width=0.85\columnwidth{}]{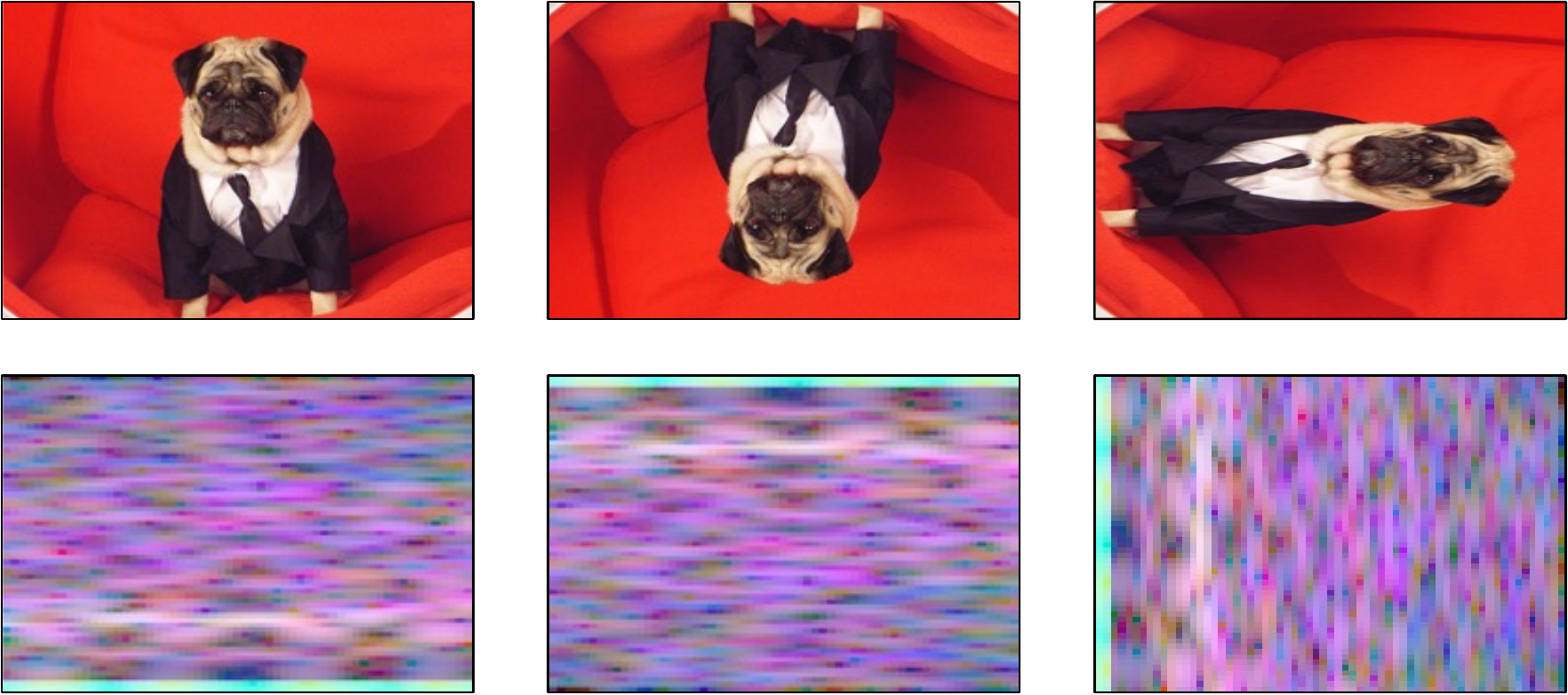}
    \setlength{\belowcaptionskip}{-8pt}
    \caption{Flipping and rotating an image from ImageNet (top) and a spectrogram image (bottom). Deformations misinterpret the spectrograms by swapping the time-frequency axes and inverting the values along an axis while preserving the label of the image from ImageNet.}
    \label{fig:incompatible}
\end{figure}

\subsection{\revised{IMU Sensing Tasks from a Vision Perspective}} \label{ssec:perceiving}

\revised{
IMU sensor data, usually presented in 1D, is often visualized in 2D to enhance understanding and analysis. 2D transformation methods~\cite{alsheikh2016deep, ravi2016deep, hur2018iss2image} have been investigated for their effectiveness in visualization and feature representations. A common approach is transforming into spectrograms, visualizing the time-frequency features for data interpretation. 

We provide an example of spectrograms from IMU sensing tasks in Figure~\ref{fig:spectrograms}, illustrating how different classes are discernible via visual features. In human activity recognition, jogging is characterized by broader spacing between the white horizontal stripes than walking, reflecting the increased frequency of motion in jogging. Similarly, in roadway classification, asphalt is represented by a darker plot than dirt, attributed to more irregular vibrations from uneven surfaces. The key to differentiating the labels is recognizing brightness, color, position, and pattern as visual information. This underlines the potential of leveraging insights from models trained on images for sensing tasks. 
In light of this observation, we explore the ability of vision models to interpret visual representations of IMU sensor data.
}

\begin{figure*}[t]
    \centering
    \includegraphics[width=0.8\textwidth{}]{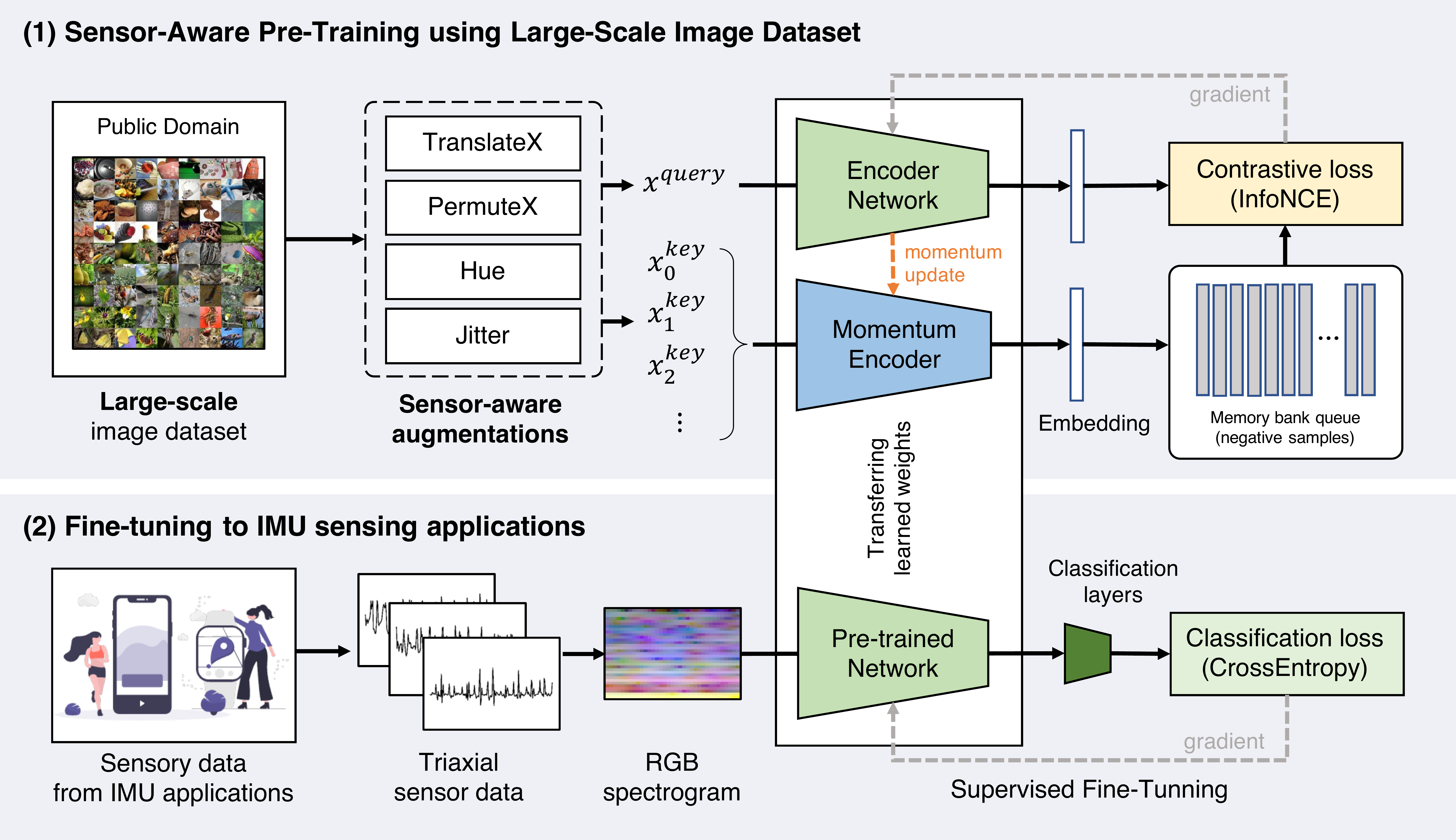}
    \setlength{\belowcaptionskip}{-14pt}
    \caption{Overview of \system{}. (1) Using the large-scale image dataset collected from the public domain, pre-training is performed via contrastive learning with specially designed sensor-aware augmentations. (2) The pre-trained model is transferred to sensing tasks using 2D-transformed triaxial IMU sensor data as input.}
    \label{fig:overview}
\end{figure*}

\subsection{\revised{Challenges in Applying Vision Knowledge to IMU Sensing}} \label{sec: challenges}

\revised{
In inspecting the potential of pre-trained models on image datasets for IMU sensing, we acknowledge the inherent uncertainties in directly applying vision knowledge. Although interpretable through basic visual cues such as brightness and patterns, sensor spectrograms possess unique properties that differ from standard images. Each axis---time and frequency---carries specific information, and their orientation indicates the scale of values. We demonstrate the potential misinterpretation of spectrograms by models trained on standard image datasets in Figure~\ref{fig:incompatible}. Unlike the standard image classification, where rotated or flipped images belong to the same class, such deformations disturb the critical information in spectrograms. Rotating swaps the time and frequency axes while flipping reverses the values in each axis. The incompatibility stemming from the unique visualization properties between traditional images and sensor spectrograms leads to potential misinterpretation and degraded performance when simply transferring common vision knowledge to IMU sensing tasks.

We tackle this challenge of tailoring image-based pre-training for models to suit IMU sensing tasks. Aiming at filtering incompatible information while reinforcing relevant insights, we propose an approach to pre-train images for IMU sensing tasks. We detail our approach in Section~\ref{sec:img2imu}.
} 



\section{\system{}} \label{sec:img2imu}

To enhance the performance of IMU sensing tasks when a fair amount of training data is difficult to obtain, we propose to utilize large-scale public image datasets to pre-train a model. 
Figure~\ref{fig:overview} overviews our \system{} that consists of two main stages: (i) pre-training a model using large-scale image datasets to learn sensor-aware knowledge through self-supervised contrastive learning, and (ii) transferring the learned knowledge from the vision model to downstream IMU sensing tasks that use 2D-transformed sensor data. 

\subsection{\revised{Converting Triaxial IMU Sensing Data to Images}} \label{section:conversion}



\begin{figure}[t]
    \centering
    \includegraphics[width=0.75\columnwidth{}]{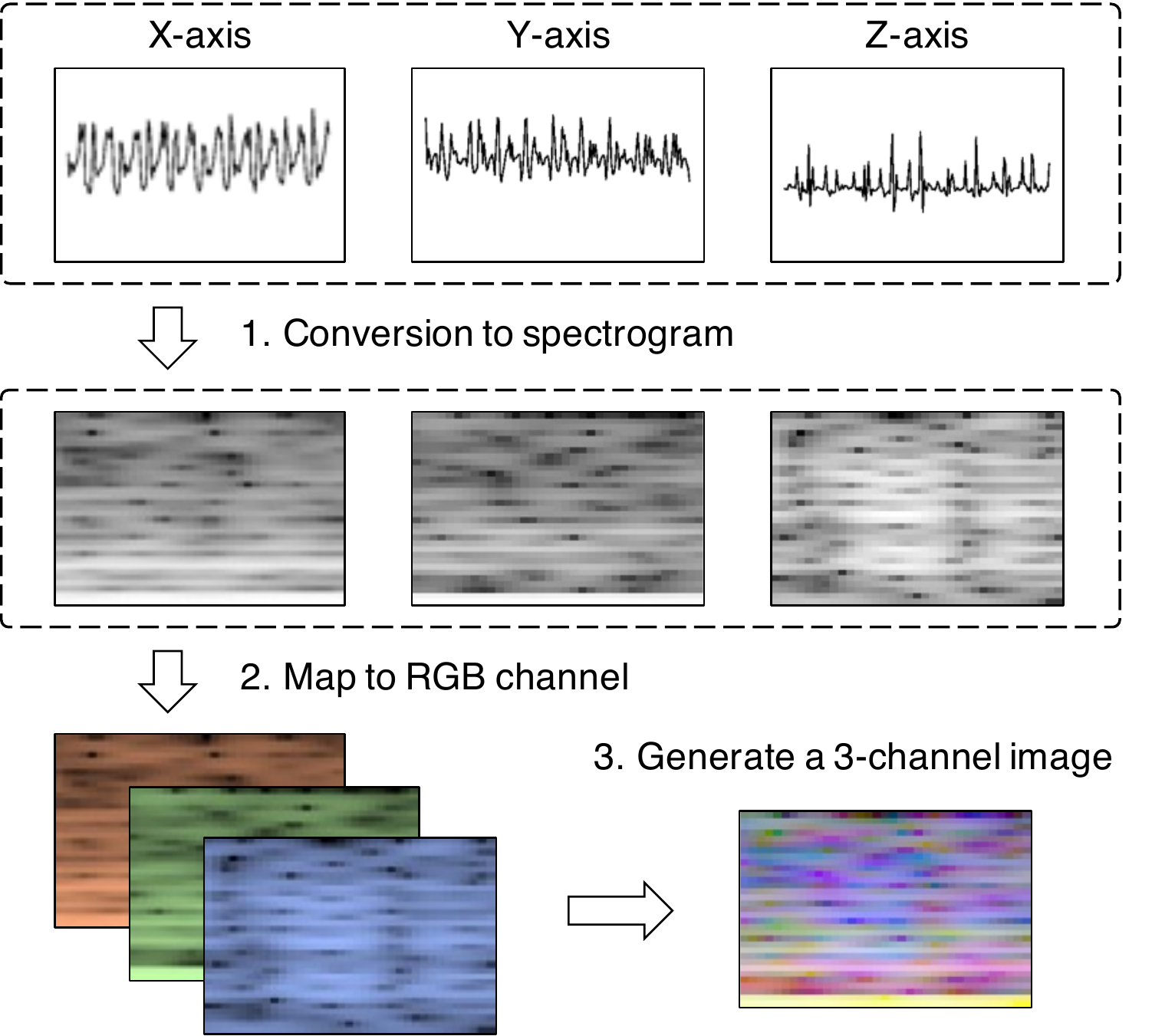}
    \setlength{\belowcaptionskip}{-10pt}
    \caption{Generation of a 3-channel 2D representation image from triaxial IMU sensing data.}
    \label{fig:conversion}
\end{figure}

\revised{Spectrograms display the intensity of frequency features along the time axis. Existing works~\cite{alsheikh2016deep, ravi2016deep, hur2018iss2image} showed that the frequency-based visualization effectively represents features for various IMU sensing tasks. Building on this foundation, we set spectrograms as our primary visualization method, expecting that the ability to interpret visual features from images can also be applied to spectrograms.

Our research scope is on applications that utilize triaxial IMU data, reflecting the common practice of measuring motion across the x, y, and z axes. To harness data in all axes for no information loss, 
we map the x, y, and z axes to the RGB channels to generate a single image, which was shown to be effective in IMU sensing tasks~\cite{alsheikh2016deep, ravi2016deep}. 
This method ensures that the intensity of motion, measured as the root mean square of the triaxial values, is reflected in the brightness, derived from the aggregation of RGB values. It also differentiates each axis's contribution through the prevalence of red, green, or blue hues. 

We acknowledge several issues in the mapping strategy. 
For instance, an effective augmentation method for sensor data is \textit{rotation}, \ie{} switching the x, y, and z axes, which is the same as changing the image's RGB color tones (\ie{} Hue). However, these RBG color tone changes would not be an ideal image augmentation method; for example, replacing the blue sky with a green sky does not make sense. This indicates that following the standard augmentation rules in the vision domain might fail to transfer knowledge to IMU sensing tasks effectively.
To handle this mismatch between sensor and image data, we propose a \textit{sensor-aware augmentation} strategy that effectively accounts for such variations, detailed in the subsequent sections.

Figure~\ref{fig:conversion} shows the overall generation process of a 3-channel 2D image from sensor data. Spectrograms are created for each channel and converted into corresponding color channels to create images. Afterward, the generated image is resized and normalized to fit the model's input requirements. We consider the spectrogram parameters, such as the number of points used in the Fast Fourier Transform (\ie{} \texttt{nfft}), as hyperparameters, which require tuning to optimize the model's performance. 

}


\subsection{\revised{Sensor-Aware Pre-Training using Image Dataset}}
\subsubsection{\revised{Contrastive Self-Supervised Learning}}

\revised{
To address the unique challenges presented by the distinct characteristics of spectrograms compared with conventional images (Figure~\ref{fig:incompatible}), \system{} employs contrastive learning~\cite{he2020momentum, chen2020simple} for pre-training. We use contrastive learning for its exceptional performance in training vast unlabeled data~\cite{chen2020big}. More importantly, it has \textit{the capability to selectively train knowledge that is valuable for IMU sensing while avoiding incompatible information from public image datasets}. 

Contrastive learning generates a pair of augmented views from a single source, ensuring that these views retain essential mutual information about their inherent characteristics. The goal during training is to enhance the model's ability to identify and align these augmented pairs while distinguishing them from unrelated examples. The model is trained to capture the intrinsic features maintained across augmentations. We focus on the strategic use of augmentations in contrastive learning; by selecting appropriate augmentations, we can direct the model to learn particular feature insights. For example, scaling augmentation teaches the model to recognize an object with different sizes as similar entities. In contrast, color augmentation trains it to understand that objects are similar with varying colors. In \system{}, we define tailored augmentations designed for IMU sensing tasks, empowering \system{} to acquire useful knowledge, detailed in Section~\ref{sec: augmentations}.
}

\system{} implements contrastive learning based on MoCo~\cite{he2020momentum} as it uses a much smaller batch size while achieving comparable performance compared with other baselines such as SimCLR~\cite{chen2020simple}. This efficiency allows operating in resource-constrained environments, resulting in greater scalability. 
Two encoders are maintained by MoCo; the query encoder and the key encoder. The query encoder generates an embedding named $q$ from a data sample. It generates embedding named positive key, $k_+$, from the positive pair of the sample, and negative keys $k_i (i=0,1,2,\ldots,K)$ that are encoded from the other data points. The training objective is to make the query $q$ distinguish the positive key~($k_+$) from the other negative keys~($k_i$). 
The query encoder is trained with InfoNCE loss~\cite{oord2018representation} during learning. We calculate the InfoNCE loss as follows: 
\begin{equation} \label{eq1}
    L_q=-\log \frac{\exp(q \cdot k_+/\tau)}{\sum_{i=0}^{K}exp(q\cdot k_i/\tau)},
\end{equation}
where $\tau$ indicates the temperature parameter for controlling the concentration level. MoCo maintains a large set of negative keys by constructing a dictionary that stores data of multiple encoded keys. A moving average is used to update the key encoder based on the weights trained from the query encoder, which enables the dictionary to be dynamic. After contrastive learning is performed on the training image data, the parameters of the query encoder network are used as pre-trained weights for the downstream IMU sensing task.


\subsubsection{\revised{IMU Sensor-Aware Augmentations}} \label{sec: augmentations}

Data augmentation preserves 
the key property of data and generates a different view of the same data. For instance, images are often rotated, flipped, and scaled to change their viewpoint while maintaining color and relative shapes. 
Using augmentations in contrastive learning, the model learns what mutual information to use to cognize the original and augmented data as the same. Augmentation types should be carefully selected based on what knowledge the model aims to acquire. The usefulness of different augmentations varies in different downstream tasks~\cite{tian2020makes}. 

Our downstream tasks take spectrograms derived from triaxial IMU sensing data as the input. Compared with the images from public datasets used for pre-training, spectrograms show unique properties. Spectrograms have \emph{directional properties along the axes}; thus, augmentations such as flipping images would damage the downstream performance as they reverse the time or frequency values. Similarly, rotating images would distort nature as \emph{each axis has fixed values of time and frequency}. Further, the RGB channels in our spectrograms indicate the triaxial axes of x, y, and z, thus we must be aware of \emph{the difference in the channel information}. These are the important domain gap between public image datasets and sensor data, and we thoughtfully select the augmentations for \system{} to bridge this gap.

\begin{figure}
\centering
    \includegraphics[height=90px]{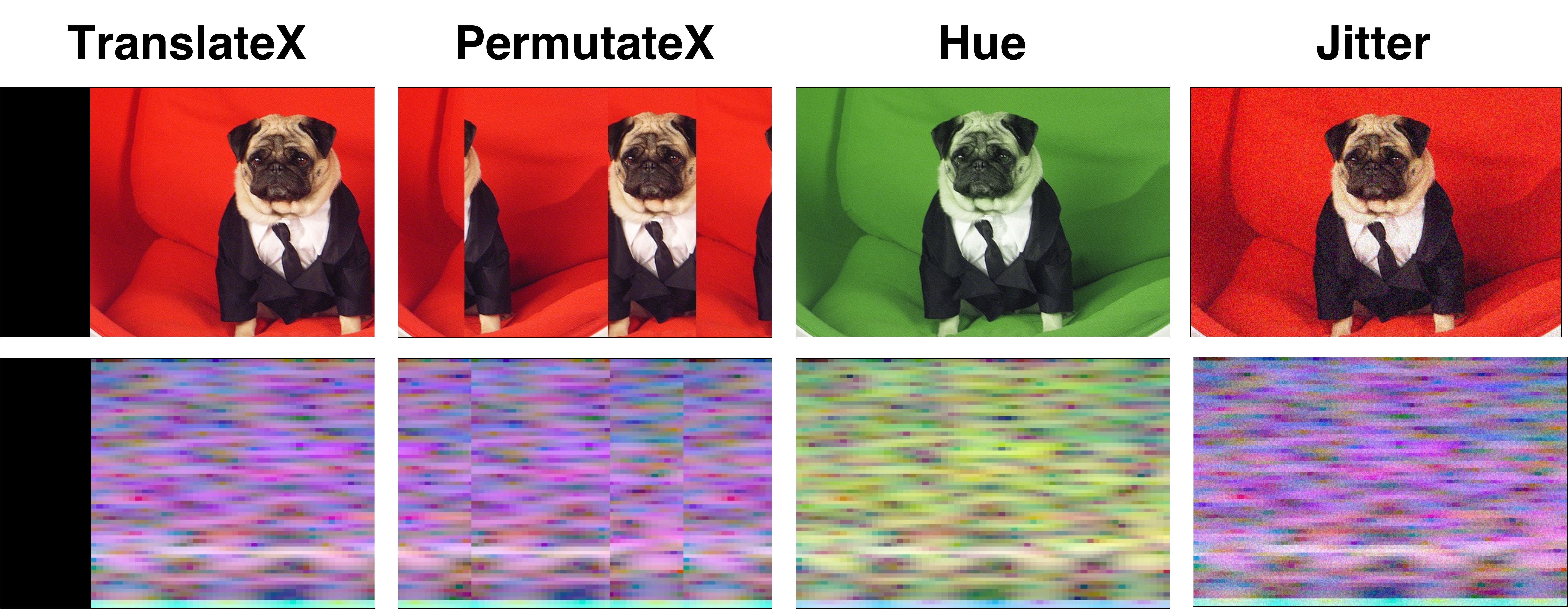}
    \setlength{\belowcaptionskip}{-5pt}
    \caption{Sensor-aware augmentations in \system{}.}
    \label{fig:augmentations_suitable}
\end{figure}

We identify the important properties of sensor data that must be preserved and define augmentations to assist the model in learning useful knowledge for downstream IMU sensing tasks. 
Figure~\ref{fig:augmentations_suitable} visualizes the selected image augmentations.

\begin{itemize}[noitemsep,topsep=5pt,leftmargin=*]
\item \textbf{\texttt{TranslateX}} randomly shifts image data on the x-axis. Sensor data are segmented into fixed-size time windows for processing. During this stage, the window can be started at any temporal point from the same context. As the key features of data are within the time window, the classification remains the same \emph{regardless of whether a window is shifted left or right over the time axis}. Based on this property, we expect that \texttt{TranslateX} benefits sensing tasks as the x-axis represents time in the spectrogram.

\item \textbf{\texttt{PermuteX}} splits data over the x-axis into multiple chunks and randomly perturbs the chunks. For sensor data, permutation is known to \emph{preserve the local temporal features while distorting the global structure of the data} to produce a different view for the same label~\cite{um2017data}. \texttt{PermuteX}  replicates the sensor augmentation by fitting the temporal perturbation into the x-axis.

\item \textbf{\texttt{Hue}} alters the color tone of image data while preserving the overall brightness and contrast. The values between RGB channels are often interchanged with \texttt{Hue}. In IMU sensing, \emph{x, y, and z channels are interchangeable based on the rotation} of the sensor. Reflecting the property, rotation is commonly used as an augmentation for triaxial sensors~\cite{um2017data}. Our approach maps the sensor data's x, y, and z channels to the RGB channel of an image. By applying \texttt{Hue}, we replicate the effect of interchangeability between the three channels in the triaxial IMU sensing data.

\item \textbf{\texttt{Jitter}} adjusts the color by adding random noise for each pixel in the image. We implemented the augmentation by injecting uniform noise centered on zero to preserve the average color information of the image. \texttt{Jitter} mimics the augmentation method of \emph{adding random noise to sensor data}. Sensors can be affected by random noise, which in turn can affect the spectrogram by making some regions brighter or darker. We adopt \texttt{Jitter} to make the model robust to the noise that could be included in sensor data from uncontrolled environments.
\end{itemize}

\begin{table*}[]
\centering
{\renewcommand{\arraystretch}{1.1}
\begin{tabularx}{\textwidth}{lCCCCCCCCC}
\Xhline{2\arrayrulewidth}
  & \textbf{original} & \multicolumn{2}{c}{\textbf{time-shifted}} & \multicolumn{2}{c}{\textbf{masked}} & \multicolumn{2}{c}{\textbf{noised}} & \multicolumn{2}{c}{\textbf{rotated}} \\ 
  \cmidrule(lr){2-2}\cmidrule(lr){3-4}\cmidrule(lr){5-6}\cmidrule(lr){7-8}\cmidrule(lr){9-10}
  & F1 & F1 & drop & F1 & drop & F1 & drop & F1 & drop \\ \hline
\texttt{T+P+H+J}~(Default) & $0.754$ & $0.545$ & $-27.75\%$ & $0.534$ & $-29.16\%$ & $0.580$ & $-23.10\%$ & $0.695$ & $-7.83\%$ \\
\texttt{P+H+J}~(w/o~\texttt{T})   & $0.686$ & \cellcolor{my_red}$0.434$ & \cellcolor{my_red}$\mathbf{-36.78\%}$ & $0.468$ & $-31.83\%$ & $0.627$ & $-8.68\%$ & $0.684$ & $-0.34\%$ \\
\texttt{T+H+J}~(w/o~\texttt{P})   & $0.687$ & $0.435$ & $\mathbf{-36.66\%}$ & \cellcolor{my_red}$0.387$ & \cellcolor{my_red}$\mathbf{-43.76\%}$ & $0.533$ & $-22.48\%$ & $0.661$ & $-3.90\%$ \\
\texttt{T+P+H}~(w/o~\texttt{J})  & $0.749$ & $0.540$ & $-27.87\%$ & $0.502$ & $-33.02\%$ & \cellcolor{my_red}$0.562$ & \cellcolor{my_red}$\mathbf{-24.91\%}$ & $0.695$ & $-7.19\%$ \\ 
\texttt{T+P+J}~(w/o~\texttt{H})  & $0.704$ & $0.559$ & $-20.68\%$ & $0.548$ & $-22.24\%$ &$ 0.62$2 & $-11.59\%$ & \cellcolor{my_red}$0.539$ & \cellcolor{my_red}$\mathbf{-23.45\%}$ \\ 
\Xhline{2\arrayrulewidth}
\end{tabularx}}
\setlength{\belowcaptionskip}{-10pt}
\caption{Evaluation showing the effect of each sensor-aware image augmentation on the robustness against the sensory augmentations applied to the WISDM~\cite{kwapisz2011activity} dataset. \texttt{T}, \texttt{P}, \texttt{H}, and \texttt{J} denotes \texttt{TranslateX}, \texttt{PermuteX}, \texttt{Hue}, and \texttt{Jitter} respectively. We report the drop of the F1-score in each sensory augmentation compared to the original data. The largest drop shown in F1-score~$(\pm1\%)$ is in bold.}
\label{table:synthetic}
\end{table*}

\subsubsection{\revised{Validation of Sensor-Aware Augmentations}}
\revised{We propose four sensor-aware image augmentations, \texttt{TranslateX}, \texttt{PermuteX}, \texttt{Hue}, and \texttt{Jitter}. We expect they improve the capacity of models to interpret key features in sensor data, which remain consistent even when sensory augmentations are applied. We investigate the impact of each sensor-aware image augmentation on the robustness of the pre-trained models, particularly against various sensory augmentations applied in sensing tasks.

We organized multiple sensor datasets, each augmented with a specific sensory augmentation. We assessed the performance of models pre-trained with our sensor-aware image augmentations when applied to the organized sensor datasets. We tested different combinations of sensor-aware image augmentations during pre-training. We monitored whether the exclusion of a specific image augmentation shows a significant performance drop when it is applied to a dataset with a particular sensory augmentation. If a decline is observed, the excluded image augmentation is crucial for the model's robustness to the sensory augmentation.

Using the WISDM~\cite{kwapisz2011activity} human activity recognition dataset as the target sensor dataset, we generated four synthetic datasets by applying distinct widely-used sensory augmentations~\cite{um2017data}. These sensory augmentations were chosen to reflect natural variability in sensor data, aligning with the principles underlying our sensor-aware image augmentations. First, we created a \textit{time-shifted} version of the data by shifting the sensor readings left or right. Second, we produced \textit{masked} data to imitate internal sensor disconnections. We adopted masking to reflect global structure distortion while preserving local temporal features instead of \textit{permutation}; due to a consistent pattern over the window, applying permutation did not significantly change the data. Next, we generated \textit{noised} data by adding uniform random noise. Finally, we created \textit{rotated} data through linear transformations that alter the axes' values interchangeably. Figure~\ref{fig:sensory_variabilities} illustrates the augmented sensor data and the resulting spectrograms.

\begin{figure}
    \centering
    \includegraphics[width=\columnwidth{}]{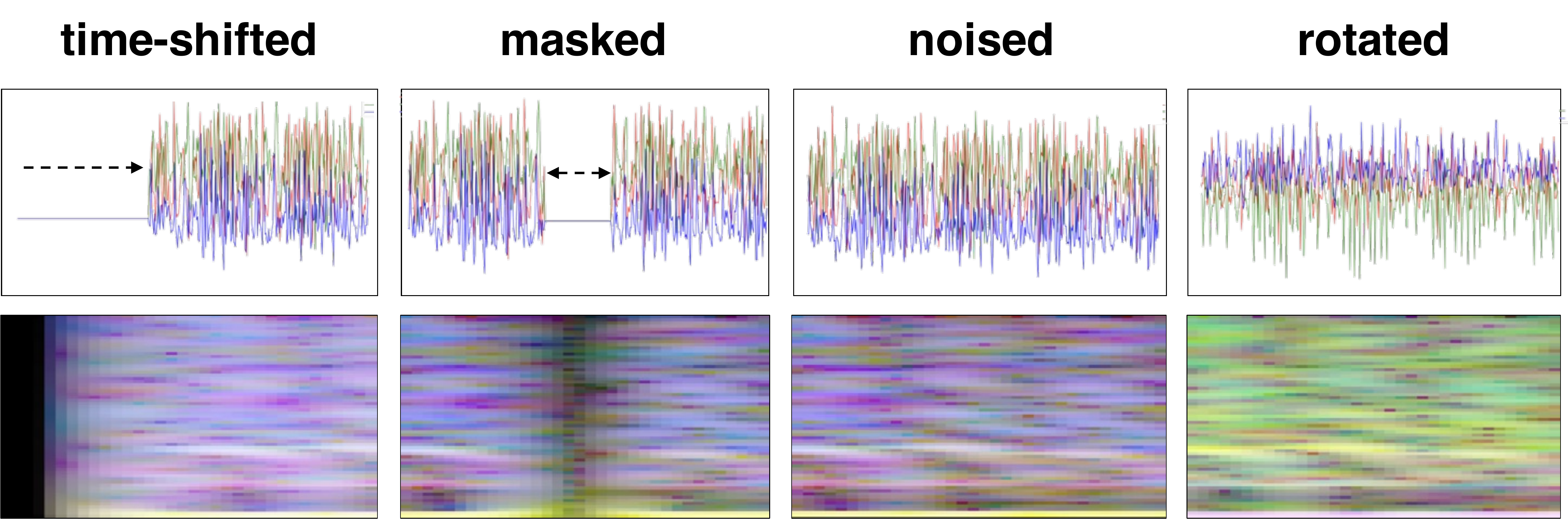}
    \setlength{\abovecaptionskip}{-6pt}
    \setlength{\belowcaptionskip}{-8pt}
    \caption{Four types of synthetic data from the WISDM~\cite{kwapisz2011activity} dataset to replicate sensor data augmentations: (i) time-shifted, (ii) masked, (iii) noised, and (iv) rotated. Both the augmented sensor data and the resulting spectrograms are shown.}
    \label{fig:sensory_variabilities}
\end{figure}

We pre-trained the models on the ImageNet~\cite{russakovsky2015imagenet} dataset with our sensor-aware image augmentations. In addition to the pre-trained model using all sensor-aware augmentations, we pre-trained four extra models. Each model was trained using three augmentations, excluding one of four augmentations per model, to observe how the exclusion impacts the model's performance on the organized datasets. 

Table~\ref{table:synthetic} shows the results for each pre-trained model across different augmented sensor datasets. We evaluated the F1-score for each model applied to each dataset. Note the performance drops when different sensory augmentations are applied compared to the original dataset. The pre-trained model with our four sensor-aware augmentations performed the best for all datasets. When we excluded each augmentation, performance dropped to different degrees. 

We examined how the absence of a specific sensor-aware image augmentation affects the robustness of the models toward the augmented datasets. First, when the data are time-shifted, the models trained without \texttt{TranslateX} and \texttt{PermuteX} showed the largest drops. This implies that \texttt{TranslateX} affects the robustness towards the time-shift of sensor data. Considering that \texttt{PermuteX} is designed to preserve the local temporal features, it also affects the performance as the robustness to time-shift requires the interpretation of local temporal features. Second, removing \texttt{PermuteX} had a noticeable effect on the masked data, which is designed to distort the global features. It verifies that \texttt{PermuteX} enhances the use of local features with the pre-trained model. With the noisy dataset, the pre-trained model without \texttt{Jitter} shows the largest drop, indicating the robustness towards the noise. Finally, with the rotated data, eliminating \texttt{Hue} from the augmentations weakens the robustness towards rotation, with significantly lower performance than others.

Through the analysis, we validated that \texttt{TranslateX}, \texttt{PermuteX}, \texttt{Hue}, and \texttt{Jitter} are sensor-aware augmentations considering the general sensory properties~\cite{um2017data}. Our study is significant not only in identifying the essential augmentations that transfer knowledge to sensing tasks but also in providing developers with the opportunity to tailor the augmentations based on the sensory properties at hand. The task-specific sensory properties must be carefully considered for the optimal augmentations use. For example, when a sensor has a fixed position, the variability in the rotation will be minor and hence \texttt{Hue} would be less effective. By adjusting the augmentations based on the correlation we revealed, developers can further optimize the performance of their applications.

}

\subsection{Fine-Tuning to IMU Sensing Tasks}

\revised{
Reflecting the scarcity of sensor data, our problem setting assumes only a few samples are available for fine-tuning. We follow a typical fine-tuning setup; the model trained on the public image dataset is fine-tuned on a small subset of data from each downstream sensing task. As shown in Figure~\ref{fig:conversion}, the data from downstream tasks, which are from IMU sensing applications, are represented as spectrograms. We adopt a popular linear evaluation protocol~\cite{wu2018unsupervised}, freezing the backbone networks and training a fully connected layer as the linear classifier at the end of the backbone network.
}

\section{Evaluation}

\subsection{Experimental Setup} \label{section:experimental-settings}

\subsubsection{Datasets}

\revised{Our approach utilizes image datasets to pre-train representations for downstream sensing tasks. We employed ImageNet~\cite{russakovsky2015imagenet}, a widely known image dataset with 1.28M samples, for pre-training \system{}. For comparison, we used the Capture-24~\cite{chan2021capture} dataset for pre-training. Capture-24 comprises accelerometer data collected from wrist-worn devices of 151 participants. It comprehensively tracks daily activities, encompassing approximately 4K hours of data sampled at 100 Hz. The Capture-24 dataset has been extensively used in previous research~\cite{haresamudram2022assessing}.

We assessed the effectiveness of the pre-trained models through their application to different downstream tasks. We set four IMU sensing tasks, all utilizing triaxial accelerometer data for classification. 
To thoroughly investigate the generalizability, we chose datasets based on the diversity of subjects, sensor position, and the nature of the tasks.

\noindent\textbf{WISDM~\cite{kwapisz2011activity}} covers human activity recognition tasks. Six different activities of sitting, standing, walking, jogging, walking downstairs, and walking upstairs, were performed by 36 participants. Participants carried smartphones in their pockets during the experiment, where accelerometer data was collected. 

\noindent\textbf{Goat Movement~\cite{kamminga2018robust}} contains activity recognition for goats on farms. Data was collected by six accelerometers 
attached to the collar-shaped device worn by five goats. Activities include stationary, walking, eating, running, and trotting. We omitted eating from our evaluation as it did not have enough samples.

\noindent\textbf{PVS~\cite{menegazzo2020multi}} is for recognizing the road features on which vehicles travel. Accelerometer devices were placed on the vehicles, and the data was measured from three drivers driving three different types of cars. We use the label information indicating the type of roadway for our main classification task: asphalt, dirt, and cobblestone. 

\noindent\textbf{Daphnet~\cite{bachlin2009wearable}} is used to detect the freeze of gait for Parkinson's disease patients. A custom wearable was attached to the patients, and the acceleration data was measured. Ten patients participated, and the wearables were positioned in three locations: ankle, leg, and waist. We use the data measured from the ankle to differentiate the positional property from WISDM.
}


\subsubsection{Data Preprocessing} \label{section:data_preprocessing}

\revised{
The ImageNet dataset was resized to $128 \times 96$ pixels. 
The images were then normalized using the statistics of ImageNet. 
The Capture-24 dataset was downsampled to 50~Hz. Given the variety of downstream classification tasks, data from Capture-24 was segmented into windows of 2, 5, and 10 seconds, each with a 50\% overlap. Separate models were pre-trained for each window size, and corresponding models were utilized for downstream tasks requiring different window sizes. The Capture-24 data was normalized using its mean and standard deviation values.

All downstream sensing data were resampled to 50~Hz to align with the pre-training frequency. Based on the specific requirements of each task, data was windowed into segments of 2, 5, or 10 seconds, using sliding windows with a 50\% overlap. The chosen window size matches the description in the respective dataset's original publication~\cite{kwapisz2011activity, kamminga2018robust, menegazzo2020multi, bachlin2009wearable}. Each dataset was divided into training, testing, and validation sets in a 60:20:20 ratio. This division was user- or subject-specific (\eg{} individual goats and vehicles). All sensory data were normalized based on the statistics of the pre-training source dataset, following 
a prior work~\cite{haresamudram2022assessing}.

For models requiring image inputs, spectrograms were generated from the sensory data. 
Spectrogram generation parameters, specifically \textit{nfft} and \textit{noverlap}, were treated as hyperparameters. A grid search was conducted to determine the optimal hyperparameters, with \textit{nfft} values $\{32, 64, 128, 256\}$ and \textit{noverlap} set at \textit{nfft} minus 2, 4, 8, and 16 for each \textit{nfft} value. As described in Section~\ref{section:conversion}, each spectrogram was concatenated into a single RGB image. These spectrogram images were resized to 128$\times$96 pixels and normalized using the ImageNet training data statistics.
}

\subsubsection{Baselines} \label{baselines}
\revised{
We compared \system{} against a total of nine baselines: four taking raw (1D) sensory data as input (\ie{} sensor-based) and five utilizing 2D-transformed spectrograms (\ie{} image-based). 

For the sensor-based baselines, we selected established self-supervised learning methods tailored to sensory data~\cite{haresamudram2022assessing}: SimCLR, Multi-task Learning, and CPC. They were pre-trained on the Capture-24 dataset~\cite{chan2021capture}, and the pre-trained weights were used for the downstream tasks that use waveform data as input. We specify the sensor-based baselines as follows. 

\begin{itemize}[leftmargin=*]
\item \noindent\textbf{Randomly-initialized (1D).} The weights of the model were randomly initialized without any pre-training. Only a few samples from each downstream task were used to train the model. It took 1D waveform data as input.
\item \noindent\textbf{SimCLR (1D)~\cite{tang2020exploring}.} We set SimCLR, specifically designed for sensory tasks, as a baseline leveraging contrastive learning~\cite{chen2020big} for sensory inputs (1D). In contrast to \system{}, this baseline applies sensory augmentations directly to the raw sensor data to generate positive views.
\item \noindent\textbf{Multi-task learning (1D)~\cite{saeed2019multi}.} It serves as a baseline for a prevalent self-supervised learning technique tailored to sensory data. It applies different sensory augmentations to create unique prediction tasks, all processed through a single encoder. By training mutual information between tasks, the encoder learns generalizable representation.
\item \noindent\textbf{Contrastive Predictive Coding (CPC) (1D)~\cite{haresamudram2022investigating}.} CPC is a self-supervised learning method that trains models to forecast future embeddings by aggregating past embeddings. This enables the model to capture the temporal dynamics in the data and adapt to sensory tasks. We used the latest version, achieving the state-of-the-art human activity recognition benchmark performance.
\end{itemize}

For the image-based baselines, we compared models pre-trained on the ImageNet~\cite{russakovsky2015imagenet} dataset, each utilizing unique pre-training strategies. These were used for downstream tasks having 2D-transformed spectrograms as input. We provide details as follows.

\begin{itemize}[leftmargin=*]
\item \noindent\textbf{Randomly-initialized (2D).} We set a baseline by randomly initializing encoder weights for image-based models. Without any pre-training, only the spectrograms of the downstream tasks are used for fine-tuning.
\item \noindent\textbf{ImageNet-supervised (2D).} We pre-trained ImageNet with supervised learning using its labels. The trained weights are transferred for the downstream tasks with spectrograms.
\item \noindent\textbf{SimSiam (2D)~\cite{chen2021exploring}.} SimSiam represents contrastive learning that bypasses the need for negative samples by using stop-gradient. It showcases the application of different loss optimization in contrastive learning. We used the augmentation set provided by the authors to implement positive views.
\item \noindent\textbf{MoCo (2D)~\cite{chen2021empirical}.} we use MoCo as a standard baseline in contrast to the sensor-aware augmentations. This approach incorporates the default augmentations provided in the latest version of MoCo: crop and resize, jittering, horizontal flipping, and Gaussian blurring.
\item \noindent\textbf{MoCo + All augmentations (2D)~\cite{shorten2019survey}.} To explore the impact of various image augmentations, this baseline uses an extensive set of image augmentations: rotating, sharpening, shearing, adjusting contrast, brightness, and color, inverting RGB values, polarizing, posterizing, equalizing, and applying automatic contrast. They were applied in the MoCo-based pre-training.
\end{itemize}
}

\newcolumntype{Y}{>{\centering\arraybackslash}X}
\begin{table*}[t]
\centering
\caption{F1-scores of \system{} and the baselines with 10 training samples per class. Sensor-based baselines, using 1D waveform input, were pre-trained on Capture-24~\cite{chan2021capture}, while image-based baselines, using 2D spectrograms, were pre-trained on ImageNet~\cite{russakovsky2015imagenet}. Encoders were frozen during fine-tuning, with only the last layer trained. Highest F1-scores in bold.}
\label{tab:eval_ssl}
{\renewcommand{\arraystretch}{1.2}
\begin{tabularx}{\textwidth}{ll*{5}{C}}
\Xhline{2\arrayrulewidth}
\multirow{2}{*}{} & \multirow{2}{*}{Pre-Training Method} & \multirow{2}{*}{WISDM} & Goat Movement & \multirow{2}{*}{PVS} & \multirow{2}{*}{Daphnet} & \cellcolor{my_blue}\multirow{2}{*}{Avg.} \\ \Xhline{2\arrayrulewidth}
\multirow{4}{*}{\shortstack[c]{Sensor-based \\methods \\(Pre-trained on \\Capture-24~\cite{chan2021capture})}}
 & Randomly-init. (1D) & 0.550\scalebox{0.7}{ $\pm$ 0.141} & 0.270\scalebox{0.7}{ $\pm$ 0.123} & 0.585\scalebox{0.7}{ $\pm$ 0.065} & 0.420\scalebox{0.7}{ $\pm$ 0.058} & \cellcolor{my_blue}0.456\scalebox{0.7}{ $\pm$ 0.184} \\
 & SimCLR~\cite{tang2020exploring} & 0.645\scalebox{0.7}{ $\pm$ 0.050} & 0.585\scalebox{0.7}{ $\pm$ 0.061} & 0.560\scalebox{0.7}{ $\pm$ 0.113} & 0.438\scalebox{0.7}{ $\pm$ 0.053} & \cellcolor{my_blue}0.557\scalebox{0.7}{ $\pm$ 0.124} \\
 & Multi-task learning~\cite{saeed2019multi} & 0.550\scalebox{0.7}{ $\pm$ 0.170} & 0.662\scalebox{0.7}{ $\pm$ 0.029} & 0.583\scalebox{0.7}{ $\pm$ 0.051} & 0.520\scalebox{0.7}{ $\pm$ 0.073} & \cellcolor{my_blue}0.579\scalebox{0.7}{ $\pm$ 0.126} \\
 & CPC~\cite{haresamudram2022investigating} & 0.552\scalebox{0.7}{ $\pm$ 0.151} & 0.650\scalebox{0.7}{ $\pm$ 0.112} & 0.578\scalebox{0.7}{ $\pm$ 0.084} & 0.517\scalebox{0.7}{ $\pm$ 0.083} & \cellcolor{my_blue}0.574\scalebox{0.7}{ $\pm$ 0.165} \\ \hline
\multirow{6}{*}{\shortstack[c]{Image-based \\methods \\(Pre-trained on \\ImageNet~\cite{russakovsky2015imagenet})}}
 & Randomly-init. (2D) & 0.374\scalebox{0.7}{ $\pm$ 0.105} & 0.314\scalebox{0.7}{ $\pm$ 0.055} & 0.483\scalebox{0.7}{ $\pm$ 0.118} & 0.456\scalebox{0.7}{ $\pm$ 0.090} & \cellcolor{my_blue}0.407\scalebox{0.7}{ $\pm$ 0.115} \\
 & ImageNet-supervised & 0.620\scalebox{0.7}{ $\pm$ 0.043} & 0.756\scalebox{0.7}{ $\pm$ 0.051} & 0.535\scalebox{0.7}{ $\pm$ 0.069} & 0.499\scalebox{0.7}{ $\pm$ 0.101} & \cellcolor{my_blue}0.603\scalebox{0.7}{ $\pm$ 0.111} \\
 & SimSiam~\cite{chen2021exploring} & 0.613\scalebox{0.7}{ $\pm$ 0.099} & 0.798\scalebox{0.7}{ $\pm$ 0.093} & 0.518\scalebox{0.7}{ $\pm$ 0.045} & 0.465\scalebox{0.7}{ $\pm$ 0.058} & \cellcolor{my_blue}0.598\scalebox{0.7}{ $\pm$ 0.143} \\
 & MoCo~\cite{chen2021empirical} & 0.689\scalebox{0.7}{ $\pm$ 0.023} & 0.801\scalebox{0.7}{ $\pm$ 0.057} & 0.569\scalebox{0.7}{ $\pm$ 0.062} & 0.502\scalebox{0.7}{ $\pm$ 0.097} & \cellcolor{my_blue}0.640\scalebox{0.7}{ $\pm$ 0.119} \\
 & MoCo + All aug.~\cite{shorten2019survey} & 0.627\scalebox{0.7}{ $\pm$ 0.035} & 0.756\scalebox{0.7}{ $\pm$ 0.061} & 0.470\scalebox{0.7}{ $\pm$ 0.071} & 0.484\scalebox{0.7}{ $\pm$ 0.093} & \cellcolor{my_blue}0.584\scalebox{0.7}{ $\pm$ 0.123} \\
\rowcolor{my_green}\cellcolor{white} & \system{} (ours) & \textbf{0.739}\scalebox{0.7}{ $\pm$ \textbf{0.038}} & \textbf{0.821}\scalebox{0.7}{ $\pm$ \textbf{0.024}} & \textbf{0.594}\scalebox{0.7}{ $\pm$ \textbf{0.053}} & \textbf{0.547}\scalebox{0.7}{ $\pm$ \textbf{0.085}} & \textbf{0.675}\scalebox{0.7}{ $\pm$ \textbf{0.114}} \\ \Xhline{2\arrayrulewidth}
\end{tabularx}}
\vspace{-8pt}
\end{table*}

\subsubsection{Training Configurations} \label{train_config}
\revised{
For \system{}, we used ResNet18~\cite{he2016deep} backbone and Adam optimizer~\cite{kingma2014adam} for both pre-training and fine-tuning. \system{} was implemented upon MoCo~\cite{he2020momentum} by replacing the augmentations to \texttt{TranslateX}, \texttt{PermuteX}, \texttt{Hue}, and \texttt{Jitter}. 
Pre-training was conducted over 40 epochs, using a learning rate of $1e^{-6}$ and a batch size of 256. We used a reduced MoCo feature dimension of 64 and a queue size of 4,096 to decrease the computational load. During fine-tuning, we loaded the pre-trained weights and replaced the last layer of ResNet18 with a randomly initialized layer. We leveraged linear evaluation protocol~\cite{wu2018unsupervised}, aiming to assess the effectiveness of the pre-trained weights as a feature extractor. The pre-trained encoder was kept frozen. Fine-tuning involved only a few samples (\eg{} 10) from each class and was conducted over 50 epochs. We adopted cosine annealing with warmup. The learning rate started from $1e^{-8}$ and increased up to $1e^{-5}$ for the initial 10 epochs and dropped to $1e^{-6}$ by the last epoch. A batch size of 4 was used for fine-tuning. 2D-transformed spectrograms were used for fine-tuning, and we conducted a grid search for optimal spectrogram generation parameters for each downstream task (\cf{} Section~\ref{section:data_preprocessing}).

All image-based baselines were built upon ResNet18. The ImageNet-supervised weights were loaded through TorchVision API. For the baselines employing MoCo, we maintained the pre-training configuration of \system{}. With SimSiam, we strictly followed the settings in its official implementation~\cite{chen2021exploring}. The fine-tuning for all image-based baselines was conducted in the same setting as \system{}.

For sensor-based baselines, we implemented 1D convolutional neural networks (CNNs) with a subsequent fully connected layer, strictly replicating the network architecture established in the prior assessment study~\cite{haresamudram2022assessing}. Specifically for CPC, we replicated the updated version~\cite{haresamudram2022investigating}, noted for its enhanced performance. Models underwent pre-training on the Capture-24 dataset for 50 epochs. Pre-training hyperparameters were refined via a grid search: learning rates from \{$1e^{-1}$, $1e^{-2}$, $1e^{-3}$, $1e^{-4}$, $1e^{-5}$\}, batch sizes from \{64, 128, 256\}---\{1024, 2048, 4096\} for SimCLR requiring larger batch, and weight decays from \{0, $1e^{-3}$, $1e^{-4}$\}. Fine-tuning mirrored the \system{} protocol, training the last layer with a frozen encoder for 50 epochs and maintaining a consistent batch size of 4. Fine-tuning hyperparameters were optimized, exploring the same range of learning rates and weight decay values as for pre-training hyperparameters.

We conducted five-fold cross-validation, segmenting the test and train sets by users or subjects (\cf{} Section~\ref{section:data_preprocessing}). Implementations utilized PyTorch and were performed on eight NVIDIA TITAN Xp GPUs.
}

\begin{figure*}[t]
    \centering
    \includegraphics[width=\textwidth{}]{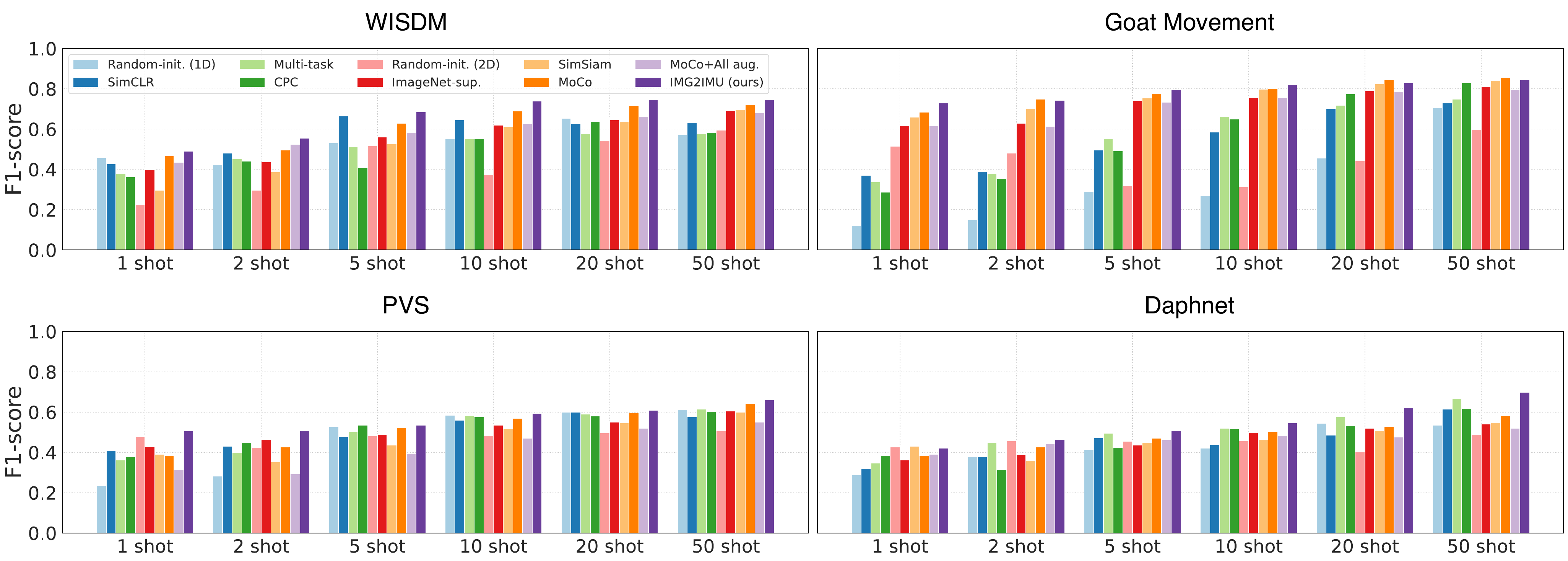}
    \setlength{\abovecaptionskip}{-8pt}
    \setlength{\belowcaptionskip}{-5pt}
    \caption{Performance of the baselines and \system{} using $n$ training samples where the number of training samples is $n\in \left \{ 1, 2, 5, 10, 20, 50 \right \}$. \system{} performs better than the baselines, particularly as n decreases.}
    \label{fig:different_samples}
\end{figure*}

\subsubsection{Metric}

The evaluation datasets contain extreme class imbalances. 
We use macro-averaged F1-score over classes as our primary performance metric, which is robust under class imbalance.

\subsection{Performance Analysis} \label{section:performance_on_sensing_tasks}

\subsubsection{Overall Results} \label{section:end-to-end}

\revised{
We conducted experiments to investigate the performance of \system{} against the baselines when only a few labeled data were available for fine-tuning toward downstream IMU sensing tasks. For all pre-trained models, we used 10 samples per class for fine-tuning. We examined the performance of the fine-tuned models on the test data of the same downstream task.

Table~\ref{tab:eval_ssl} shows the result, where \system{} consistently demonstrates superior performance over all baselines. When compared to sensor-based baselines, \system{} achieves a significant improvement, surpassing the highest F1-score by 9.8\%p. This performance of \system{} is not simply attributed to the adoption of 2D-transformed inputs, as evidenced by the poor average F1-score (0.407) of randomly initialized models with 2D inputs compared with the F1-score of those with 1D sensory inputs (0.456). This highlights the efficacy of \system{}'s pre-training, which yielded a substantial F1-score increase from 0.407 to 0.675. This is a marked contrast to the modest gain of the sensor-based pre-training, which increased at most from 0.456 to 0.579. This result indicates that pre-training using Capture-24 is limited in being applied across downstream tasks involving heterogeneous sensor positions, subjects, or task types. In contrast, \system{} shows that pre-training with the ImageNet dataset---despite its lack of spectrogram images---enables the model to interpret visual features within spectrograms, illustrating better applicability of \system{} in various sensory tasks.

Comparison with image-based baselines shows the effectiveness of \system{} pre-training, as they all use the same ImageNet dataset. \system{} surpasses pre-training using supervised learning and SimSiam by a margin greater than 7\%p. Comparison with two MoCo-based baselines underscores the impact of augmentation selection in \system{}. Despite the default MoCo augmentations achieving the highest performance for typical vision benchmarks (\eg{} CIFAR-100~\cite{krizhevsky2009learning}), our findings indicate that \system{}'s specific augmentations are more appropriate for sensory tasks (0.640$\rightarrow$0.675). Furthermore, comparison with MoCo + All augmentations~\cite{shorten2019survey} (0.584$\rightarrow$0.675) suggests that merely increasing the augmentations does not guarantee enhanced performance. This implies the effect of augmentation selection to transfer knowledge from image data to sensor data, proving that our strategy is well-suited for sensing tasks.

Additional experiments were conducted by changing the number of samples used for training (\{1, 2, 5, 10, 20, 50\} per class). Figure~\ref{fig:different_samples} shows that generally \system{} performs better than the baselines,  
especially when training data is limited, across diverse sensing tasks. Note that we do not limit the potential of \system{} to be trained solely with ImageNet. We anticipate using larger datasets such as LAION-5B would result in greater benefits.
}

\subsubsection{Visualizing Semantic Class-Discriminative Heatmaps} \label{section:visualization}

\begin{figure}[t]
    \centering
    \includegraphics[width=1\columnwidth{}]{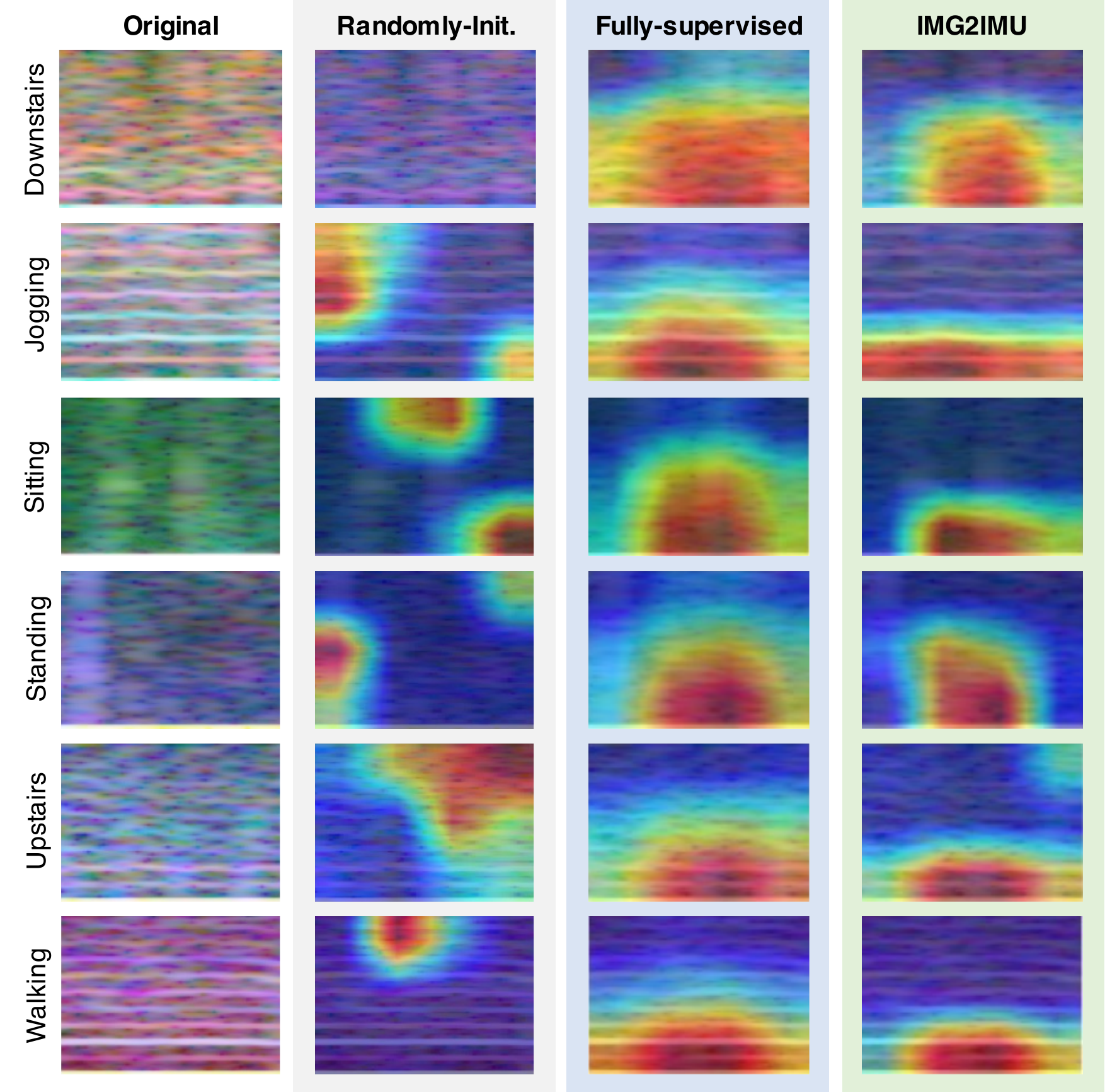}
    \setlength{\abovecaptionskip}{-0pt}
    \setlength{\belowcaptionskip}{-10pt}
    \caption{Grad-CAM comparison among randomly-initialized, fully-supervised, and \system{} models. Highlighted areas in red indicate the part where the model focused on.}
    \label{fig:gradcam}
\end{figure}

\revised{
To further understand whether \system{} works with sensory information, we examined \textit{how similar the representation learned from images is to that learned under the supervision of sufficient sensor data}. We utilized Grad-CAM~\cite{selvaraju2017grad} 
to visualize the feature interpretation of the \system{} pre-trained model. By tracking the gradient flows in convolutional layers, Grad-CAM visualizes a class-discriminative localization map highlighting influential regions in images that contribute to predicting the target concept. 
We compared \system{} against a fully-supervised model that is trained with the full training data of WISDM. We set a randomly initialized model as a baseline to show the default heatmap from an image-based model without any pre-training. We kept the convolutional layers of \system{} frozen to preserve the weights trained from ImageNet. We applied Grad-CAM on the last convolutional layer to obtain spatial information.

Figure~\ref{fig:gradcam} depicts the heatmaps drawn by Grad-CAM. 
We randomly selected a sample from each class in the WISDM dataset. \system{} and fully-supervised models spotlight analogous areas in the spectrograms for all class pairs. While the layers in \system{} are trained only on the public image dataset, the Grad-CAM results indicate that the model properly interprets sensor data, following the ideal model trained with full supervision of sensor data. Overall, the low-frequency band is emphasized in the spectrogram. A wide range of temporal features is highlighted for activities lasting long periods, such as walking and jogging, whereas a narrow range of features is underlined for upstairs and downstairs. We believe this result suggests that the representation learned from the public image dataset in \system{} applies to sensing tasks.
}

\subsection{\revised{On-device Computational Overhead}} \label{section:computation-overhead}
\revised{
The significance of on-device machine learning systems 
mainly lies in the protection of user data and the decentralization of computations. 
We consider an on-device deployment scenario where we evaluate \system{}'s real-time operation capabilities. We assume that pre-training and fine-tuning are completed with a powerful server, after which the model is deployed to a device. Consequently, our focus is on evaluating the overhead associated with on-device inference. 

Our framework incurs overhead from the transformation of sensor data into spectrograms and the use of 2D network architecture. To quantify the overhead, we implemented the \system{} inference framework on Android smartphones using the PyTorch java library. We used three commodity smartphones~(Galaxy S22 Ultra, Galaxy S20 Ultra, and Pixel 2XL) and the WISDM dataset with the fine-tuned \system{}. The average execution time was measured as the overhead by conducting ten experiments. 

The results indicate that generating the RGB spectrogram consumed an average of 48.72~ms, and the inference for the generated spectrogram took an average of 16.5~ms on Galaxy S22 Ultra. It took 55.33~ms and 26.47~ms on Galaxy S20 Ultra, 88.86~ms and 40.67~ms on Pixel 2XL. Overall, the end-to-end computation time from the framework was less than 0.2 seconds, which illustrates \system{}'s overhead is negligible to run on-device real-time inference. 
}

\section{Discussion}

\subsection{Potential for Exploring Sensor-Aware Augmentations}

The selection of augmentation types in contrastive learning strongly impacts the performance of downstream tasks. \system{} defines four augmentations that benefit contrastive learning for IMU sensing tasks. This augmentation design was derived from the key invariants 
in sensing applications, referring to the widely accepted sensor data augmentations~\cite{um2017data}. While we also attempted other types of image augmentation, such as \texttt{Brightness} and \texttt{Contrast}, 
they did not show clear correlations. Nevertheless, as there are numerous invariants in sensor data, there could be other augmentations useful for sensing applications. 
More augmentations could be built upon and potentially further improve the pre-trained model's performance with \system{}.




\subsection{\revised{Optimizing 2D Transformation Process of Sensor Data}}

To apply the knowledge learned from images, we transform the IMU sensor data into spectrogram images. This design choice was based on the fact that spectrogram is a widely accepted 2D-transformation technique for sensors. While we showed that conversion to spectrograms could benefit a diverse range of sensing tasks when combined with \system{}, there could be occasions when spectrograms fail to capture important features. For instance, the spectrogram fails to reflect the data characteristics when the Fourier transform is performed using an \emph{nfft} parameter that is too large or too small in our conversion process. In other words, the spectrogram conversion process is sensitive to a few parameters. To address this issue, other types of sensor 2D-transformation methods could be additionally incorporated. It was reported that other types of 2D representations~\cite{hur2018iss2image} for sensor data work well as input features for sensory classification tasks. The 2D representations could be used with \system{} by designing new types of augmentations that are suitable for their conversion method.

\section{Related Work}

\subsection{Self-Supervised Learning for Sensing}

Prior works~\cite{saeed2019multi, yuan2022self} applied self-supervised learning using a multi-task Transformation Prediction Network~(TPN) for human activity recognition~(HAR). Using TPN, the original data is augmented with a random augmentation, and the network is trained to predict the type of augmentation applied. SelfHAR~\cite{tang2021selfhar} integrated the ideas of multi-task learning and teacher-student self-learning to create an effective semi-supervised learning framework. 

Contrastive learning is another effective method where MoCo~\cite{he2020momentum, chen2021empirical} and SimCLR~\cite{chen2020simple, chen2020big} are representative frameworks. They have been redesigned for HAR as MoCoHAR~\cite{wang2021sensor}, SimCLR for HAR~\cite{tang2020exploring}, and CSSHAR~\cite{khaertdinov2021contrastive}. Several studies~\cite{haresamudram2021contrastive, haresamudram2022investigating} adopted Contrastive Predictive Coding (CPC), which trains an encoder to predict the next sequence chunk based on previous sequences. 

Masked region reconstruction~\cite{haresamudram2020masked, xu2021limu} is also adopted as a self-supervised learning strategy for sensory data. Haresaumudram, et al.~\cite{haresamudram2022assessing} conducted an assessment of seven state-of-the-art self-supervised learning methods applied to HAR, including BYOL~\cite{grill2020bootstrap} and SimSiam~\cite{chen2021exploring} in addition to previously discussed methods. 

While these studies showed their effectiveness for HAR tasks, IMU sensing applications include diverse target tasks~\cite{carlos2019smartphone, ismail2019review}, target subjects~\cite{kamminga2018robust}, and data collection protocols~\cite{shin2022mydj}. As publically available large-scale sensor datasets~\cite{chan2021capture} are centered on HAR but lack such diversities, the pre-trained model for sensing has poor generalizability to various sensing tasks (\cf{} Section~\ref{section:performance_on_sensing_tasks}). \system{} resolve this challenge by interpreting IMU sensor data as images and utilizing models pre-trained from a larger scale of vision data.

\subsection{Use of Cross-Modal Data for Sensing}

To enhance self-supervised learning for IMU sensing, learning with data from different modalities has been proposed. ColloSSL~\cite{jain2022collossl} and COCOA~\cite{deldari2022cocoa} used sensor data in cross-modal as positive view pairs for contrastive learning. Vision2Sensor~\cite{radu2019vision2sensor} proposed an approach of vision-to-sensor label transmission to learn through the labels generated by vision-based activity recognition. However, these approaches require careful synchronization between the different modalities. \system{}, on the other hand, eliminates the need for data synchronization as pre-training and fine-tuning datasets are learned independently. IMU2Doppler~\cite{bhalla2021imu2doppler} adopted domain adaptation to utilize data from IMU sensors to train a model that can work as a base of activity recognition using mmWave. 

\section{Conclusion}

We presented \system{} that utilizes the learned representation from images to IMU sensing tasks. We proposed a new contrastive learning method that employs image augmentations explicitly designed for sensing applications and correlates each augmentation type with sensory properties. Our evaluations demonstrated that \system{} improves performance on a variety of IMU sensing applications when fine-tuned to the learned representations. \system{} showcased how vision knowledge can be effectively translated to IMU sensing tasks and is beneficial for IMU sensing applications that lack large-scale training data. 


 
%

\bibliographystyle{IEEEtran}
\bibliography{references}

\begin{thebibliography}{10}
\providecommand{\url}[1]{#1}
\csname url@samestyle\endcsname
\providecommand{\newblock}{\relax}
\providecommand{\bibinfo}[2]{#2}
\providecommand{\BIBentrySTDinterwordspacing}{\spaceskip=0pt\relax}
\providecommand{\BIBentryALTinterwordstretchfactor}{4}
\providecommand{\BIBentryALTinterwordspacing}{\spaceskip=\fontdimen2\font plus
\BIBentryALTinterwordstretchfactor\fontdimen3\font minus \fontdimen4\font\relax}
\providecommand{\BIBforeignlanguage}[2]{{%
\expandafter\ifx\csname l@#1\endcsname\relax
\typeout{** WARNING: IEEEtran.bst: No hyphenation pattern has been}%
\typeout{** loaded for the language `#1'. Using the pattern for}%
\typeout{** the default language instead.}%
\else
\language=\csname l@#1\endcsname
\fi
#2}}
\providecommand{\BIBdecl}{\relax}
\BIBdecl

\bibitem{kwapisz2011activity}
J.~R. Kwapisz, G.~M. Weiss, and S.~A. Moore, ``Activity recognition using cell phone accelerometers,'' \emph{ACM SigKDD Explorations Newsletter}, vol.~12, no.~2, pp. 74--82, 2011.

\bibitem{chavarriaga2013opportunity}
R.~Chavarriaga, H.~Sagha, A.~Calatroni, S.~T. Digumarti, G.~Tr{\"o}ster, J.~d.~R. Mill{\'a}n, and D.~Roggen, ``The opportunity challenge: A benchmark database for on-body sensor-based activity recognition,'' \emph{Pattern Recognition Letters}, vol.~34, no.~15, pp. 2033--2042, 2013.

\bibitem{carlos2019smartphone}
M.~R. Carlos, L.~C. Gonz{\'a}lez, J.~Wahlstr{\"o}m, G.~Ram{\'\i}rez, F.~Mart{\'\i}nez, and G.~Runger, ``How smartphone accelerometers reveal aggressive driving behavior?—the key is the representation,'' \emph{IEEE Transactions on Intelligent Transportation Systems}, vol.~21, no.~8, pp. 3377--3387, 2019.

\bibitem{gonzalez2017learning}
L.~C. Gonz{\'a}lez, R.~Moreno, H.~J. Escalante, F.~Mart{\'\i}nez, and M.~R. Carlos, ``Learning roadway surface disruption patterns using the bag of words representation,'' \emph{IEEE Transactions on Intelligent Transportation Systems}, vol.~18, no.~11, pp. 2916--2928, 2017.

\bibitem{kamminga2018robust}
J.~W. Kamminga, D.~V. Le, J.~P. Meijers, H.~Bisby, N.~Meratnia, and P.~J. Havinga, ``Robust sensor-orientation-independent feature selection for animal activity recognition on collar tags,'' \emph{Proceedings of the ACM on Interactive, Mobile, Wearable and Ubiquitous Technologies}, vol.~2, no.~1, pp. 1--27, 2018.

\bibitem{ismail2019review}
M.~I.~M. Ismail, R.~A. Dziyauddin, N.~A.~A. Salleh, F.~Muhammad-Sukki, N.~A. Bani, M.~A.~M. Izhar, and L.~A. Latiff, ``A review of vibration detection methods using accelerometer sensors for water pipeline leakage,'' \emph{IEEE access}, vol.~7, pp. 51\,965--51\,981, 2019.

\bibitem{kwon2011validation}
S.~Kwon, J.~Lee, G.~S. Chung, and K.~S. Park, ``Validation of heart rate extraction through an iphone accelerometer,'' in \emph{2011 Annual International Conference of the IEEE Engineering in Medicine and Biology Society}.\hskip 1em plus 0.5em minus 0.4em\relax IEEE, 2011, pp. 5260--5263.

\bibitem{bengio2013representation}
Y.~Bengio, A.~Courville, and P.~Vincent, ``Representation learning: A review and new perspectives,'' \emph{IEEE transactions on pattern analysis and machine intelligence}, vol.~35, no.~8, pp. 1798--1828, 2013.

\bibitem{devlin2018bert}
J.~Devlin, M.-W. Chang, K.~Lee, and K.~Toutanova, ``Bert: Pre-training of deep bidirectional transformers for language understanding,'' \emph{arXiv preprint arXiv:1810.04805}, 2018.

\bibitem{chowdhery2023palm}
A.~Chowdhery, S.~Narang, J.~Devlin, M.~Bosma, G.~Mishra, A.~Roberts, P.~Barham, H.~W. Chung, C.~Sutton, S.~Gehrmann \emph{et~al.}, ``Palm: Scaling language modeling with pathways,'' \emph{Journal of Machine Learning Research}, vol.~24, no. 240, pp. 1--113, 2023.

\bibitem{brown2020language}
T.~Brown, B.~Mann, N.~Ryder, M.~Subbiah, J.~D. Kaplan, P.~Dhariwal, A.~Neelakantan, P.~Shyam, G.~Sastry, A.~Askell \emph{et~al.}, ``Language models are few-shot learners,'' \emph{Advances in neural information processing systems}, vol.~33, pp. 1877--1901, 2020.

\bibitem{russakovsky2015imagenet}
O.~Russakovsky, J.~Deng, H.~Su, J.~Krause, S.~Satheesh, S.~Ma, Z.~Huang, A.~Karpathy, A.~Khosla, M.~Bernstein \emph{et~al.}, ``Imagenet large scale visual recognition challenge,'' \emph{International journal of computer vision}, vol. 115, no.~3, pp. 211--252, 2015.

\bibitem{lin2014microsoft}
T.-Y. Lin, M.~Maire, S.~Belongie, J.~Hays, P.~Perona, D.~Ramanan, P.~Doll{\'a}r, and C.~L. Zitnick, ``Microsoft coco: Common objects in context,'' in \emph{European conference on computer vision}.\hskip 1em plus 0.5em minus 0.4em\relax Springer, 2014, pp. 740--755.

\bibitem{schuhmann2022laion}
C.~Schuhmann, R.~Beaumont, R.~Vencu, C.~Gordon, R.~Wightman, M.~Cherti, T.~Coombes, A.~Katta, C.~Mullis, M.~Wortsman \emph{et~al.}, ``Laion-5b: An open large-scale dataset for training next generation image-text models,'' \emph{arXiv preprint arXiv:2210.08402}, 2022.

\bibitem{chen2020big}
T.~Chen, S.~Kornblith, K.~Swersky, M.~Norouzi, and G.~E. Hinton, ``Big self-supervised models are strong semi-supervised learners,'' \emph{Advances in neural information processing systems}, vol.~33, pp. 22\,243--22\,255, 2020.

\bibitem{haresamudram2022assessing}
H.~Haresamudram, I.~Essa, and T.~Pl{\"o}tz, ``Assessing the state of self-supervised human activity recognition using wearables,'' \emph{Proceedings of the ACM on Interactive, Mobile, Wearable and Ubiquitous Technologies}, vol.~6, no.~3, pp. 1--47, 2022.

\bibitem{chan2021capture}
S.~Chan~Chang, R.~Walmsley, J.~Gershuny, T.~Harms, E.~Thomas, K.~Milton, P.~Kelly, C.~Foster, A.~Wong, N.~Gray \emph{et~al.}, ``Capture-24: Activity tracker dataset for human activity recognition,'' 2021.

\bibitem{gershuny2020testing}
J.~Gershuny, T.~Harms, A.~Doherty, E.~Thomas, K.~Milton, P.~Kelly, and C.~Foster, ``Testing self-report time-use diaries against objective instruments in real time,'' \emph{Sociological Methodology}, vol.~50, no.~1, pp. 318--349, 2020.

\bibitem{willetts2018statistical}
M.~Willetts, S.~Hollowell, L.~Aslett, C.~Holmes, and A.~Doherty, ``Statistical machine learning of sleep and physical activity phenotypes from sensor data in 96,220 uk biobank participants,'' \emph{Scientific reports}, vol.~8, no.~1, p. 7961, 2018.

\bibitem{doherty2017large}
A.~Doherty, D.~Jackson, N.~Hammerla, T.~Pl{\"o}tz, P.~Olivier, M.~H. Granat, T.~White, V.~T. Van~Hees, M.~I. Trenell, C.~G. Owen \emph{et~al.}, ``Large scale population assessment of physical activity using wrist worn accelerometers: the uk biobank study,'' \emph{PloS one}, vol.~12, no.~2, p. e0169649, 2017.

\bibitem{jiang2015human}
W.~Jiang and Z.~Yin, ``Human activity recognition using wearable sensors by deep convolutional neural networks,'' in \emph{Proceedings of the 23rd ACM international conference on Multimedia}, 2015, pp. 1307--1310.

\bibitem{alsheikh2016deep}
M.~A. Alsheikh, A.~Selim, D.~Niyato, L.~Doyle, S.~Lin, and H.-P. Tan, ``Deep activity recognition models with triaxial accelerometers,'' in \emph{Workshops at the Thirtieth AAAI Conference on Artificial Intelligence}, 2016.

\bibitem{hur2018iss2image}
T.~Hur, J.~Bang, T.~Huynh-The, J.~Lee, J.-I. Kim, and S.~Lee, ``Iss2image: A novel signal-encoding technique for cnn-based human activity recognition,'' \emph{Sensors}, vol.~18, no.~11, p. 3910, 2018.

\bibitem{bachlin2009wearable}
M.~Bachlin, M.~Plotnik, D.~Roggen, I.~Maidan, J.~M. Hausdorff, N.~Giladi, and G.~Troster, ``Wearable assistant for parkinson’s disease patients with the freezing of gait symptom,'' \emph{IEEE Transactions on Information Technology in Biomedicine}, vol.~14, no.~2, pp. 436--446, 2009.

\bibitem{casilari2017umafall}
E.~Casilari, J.~A. Santoyo-Ram{\'o}n, and J.~M. Cano-Garc{\'\i}a, ``Umafall: A multisensor dataset for the research on automatic fall detection,'' \emph{Procedia Computer Science}, vol. 110, pp. 32--39, 2017.

\bibitem{stromback2020mm}
D.~Str{\"o}mb{\"a}ck, S.~Huang, and V.~Radu, ``Mm-fit: Multimodal deep learning for automatic exercise logging across sensing devices,'' \emph{Proceedings of the ACM on Interactive, Mobile, Wearable and Ubiquitous Technologies}, vol.~4, no.~4, pp. 1--22, 2020.

\bibitem{brunner2019swimming}
G.~Brunner, D.~Melnyk, B.~Sigf{\'u}sson, and R.~Wattenhofer, ``Swimming style recognition and lap counting using a smartwatch and deep learning,'' in \emph{Proceedings of the 2019 ACM International Symposium on Wearable Computers}, 2019, pp. 23--31.

\bibitem{menegazzo2020multi}
J.~Menegazzo and A.~von Wangenheim, ``Multi-contextual and multi-aspect analysis for road surface type classification through inertial sensors and deep learning,'' in \emph{2020 X Brazilian Symposium on Computing Systems Engineering (SBESC)}.\hskip 1em plus 0.5em minus 0.4em\relax IEEE, 2020, pp. 1--8.

\bibitem{dosovitskiy2020image}
A.~Dosovitskiy, L.~Beyer, A.~Kolesnikov, D.~Weissenborn, X.~Zhai, T.~Unterthiner, M.~Dehghani, M.~Minderer, G.~Heigold, S.~Gelly \emph{et~al.}, ``An image is worth 16x16 words: Transformers for image recognition at scale,'' \emph{arXiv preprint arXiv:2010.11929}, 2020.

\bibitem{kolesnikov2020big}
A.~Kolesnikov, L.~Beyer, X.~Zhai, J.~Puigcerver, J.~Yung, S.~Gelly, and N.~Houlsby, ``Big transfer (bit): General visual representation learning,'' in \emph{Computer Vision--ECCV 2020: 16th European Conference, Glasgow, UK, August 23--28, 2020, Proceedings, Part V 16}.\hskip 1em plus 0.5em minus 0.4em\relax Springer, 2020, pp. 491--507.

\bibitem{zhai2022scaling}
X.~Zhai, A.~Kolesnikov, N.~Houlsby, and L.~Beyer, ``Scaling vision transformers,'' in \emph{Proceedings of the IEEE/CVF Conference on Computer Vision and Pattern Recognition}, 2022, pp. 12\,104--12\,113.

\bibitem{changpinyo2021conceptual}
S.~Changpinyo, P.~Sharma, N.~Ding, and R.~Soricut, ``Conceptual 12m: Pushing web-scale image-text pre-training to recognize long-tail visual concepts,'' in \emph{Proceedings of the IEEE/CVF Conference on Computer Vision and Pattern Recognition}, 2021, pp. 3558--3568.

\bibitem{thomee2016yfcc100m}
B.~Thomee, D.~A. Shamma, G.~Friedland, B.~Elizalde, K.~Ni, D.~Poland, D.~Borth, and L.-J. Li, ``Yfcc100m: The new data in multimedia research,'' \emph{Communications of the ACM}, vol.~59, no.~2, pp. 64--73, 2016.

\bibitem{yalniz2019billion}
I.~Z. Yalniz, H.~J{\'e}gou, K.~Chen, M.~Paluri, and D.~Mahajan, ``Billion-scale semi-supervised learning for image classification,'' \emph{arXiv preprint arXiv:1905.00546}, 2019.

\bibitem{azizi2021big}
S.~Azizi, B.~Mustafa, F.~Ryan, Z.~Beaver, J.~Freyberg, J.~Deaton, A.~Loh, A.~Karthikesalingam, S.~Kornblith, T.~Chen \emph{et~al.}, ``Big self-supervised models advance medical image classification,'' in \emph{Proceedings of the IEEE/CVF International Conference on Computer Vision}, 2021, pp. 3478--3488.

\bibitem{shin2021self}
S.~Shin, J.~Kim, Y.~Yu, S.~Lee, and K.~Lee, ``Self-supervised transfer learning from natural images for sound classification,'' \emph{Applied Sciences}, vol.~11, no.~7, p. 3043, 2021.

\bibitem{ravi2016deep}
D.~Ravi, C.~Wong, B.~Lo, and G.-Z. Yang, ``Deep learning for human activity recognition: A resource efficient implementation on low-power devices,'' in \emph{2016 IEEE 13th international conference on wearable and implantable body sensor networks (BSN)}.\hskip 1em plus 0.5em minus 0.4em\relax IEEE, 2016, pp. 71--76.

\bibitem{he2020momentum}
K.~He, H.~Fan, Y.~Wu, S.~Xie, and R.~Girshick, ``Momentum contrast for unsupervised visual representation learning,'' in \emph{Proceedings of the IEEE/CVF conference on computer vision and pattern recognition}, 2020, pp. 9729--9738.

\bibitem{chen2020simple}
T.~Chen, S.~Kornblith, M.~Norouzi, and G.~Hinton, ``A simple framework for contrastive learning of visual representations,'' in \emph{International conference on machine learning}.\hskip 1em plus 0.5em minus 0.4em\relax PMLR, 2020, pp. 1597--1607.

\bibitem{oord2018representation}
A.~v.~d. Oord, Y.~Li, and O.~Vinyals, ``Representation learning with contrastive predictive coding,'' \emph{arXiv preprint arXiv:1807.03748}, 2018.

\bibitem{tian2020makes}
Y.~Tian, C.~Sun, B.~Poole, D.~Krishnan, C.~Schmid, and P.~Isola, ``What makes for good views for contrastive learning?'' \emph{Advances in Neural Information Processing Systems}, vol.~33, pp. 6827--6839, 2020.

\bibitem{um2017data}
T.~T. Um, F.~M. Pfister, D.~Pichler, S.~Endo, M.~Lang, S.~Hirche, U.~Fietzek, and D.~Kuli{\'c}, ``Data augmentation of wearable sensor data for parkinson’s disease monitoring using convolutional neural networks,'' in \emph{Proceedings of the 19th ACM international conference on multimodal interaction}, 2017, pp. 216--220.

\bibitem{wu2018unsupervised}
Z.~Wu, Y.~Xiong, S.~X. Yu, and D.~Lin, ``Unsupervised feature learning via non-parametric instance discrimination,'' in \emph{Proceedings of the IEEE conference on computer vision and pattern recognition}, 2018, pp. 3733--3742.

\bibitem{tang2020exploring}
C.~I. Tang, I.~Perez-Pozuelo, D.~Spathis, and C.~Mascolo, ``Exploring contrastive learning in human activity recognition for healthcare,'' \emph{arXiv preprint arXiv:2011.11542}, 2020.

\bibitem{saeed2019multi}
A.~Saeed, T.~Ozcelebi, and J.~Lukkien, ``Multi-task self-supervised learning for human activity detection,'' \emph{Proceedings of the ACM on Interactive, Mobile, Wearable and Ubiquitous Technologies}, vol.~3, no.~2, pp. 1--30, 2019.

\bibitem{haresamudram2022investigating}
H.~Haresamudram, I.~Essa, and T.~Ploetz, ``Investigating enhancements to contrastive predictive coding for human activity recognition,'' \emph{arXiv preprint arXiv:2211.06173}, 2022.

\bibitem{chen2021exploring}
X.~Chen and K.~He, ``Exploring simple siamese representation learning,'' in \emph{Proceedings of the IEEE/CVF conference on computer vision and pattern recognition}, 2021, pp. 15\,750--15\,758.

\bibitem{chen2021empirical}
X.~Chen, S.~Xie, and K.~He, ``An empirical study of training self-supervised vision transformers,'' in \emph{Proceedings of the IEEE/CVF International Conference on Computer Vision}, 2021, pp. 9640--9649.

\bibitem{shorten2019survey}
C.~Shorten and T.~M. Khoshgoftaar, ``A survey on image data augmentation for deep learning,'' \emph{Journal of big data}, vol.~6, no.~1, pp. 1--48, 2019.

\bibitem{he2016deep}
K.~He, X.~Zhang, S.~Ren, and J.~Sun, ``Deep residual learning for image recognition,'' in \emph{Proceedings of the IEEE conference on computer vision and pattern recognition}, 2016, pp. 770--778.

\bibitem{kingma2014adam}
D.~P. Kingma and J.~Ba, ``Adam: A method for stochastic optimization,'' \emph{arXiv preprint arXiv:1412.6980}, 2014.

\bibitem{krizhevsky2009learning}
A.~Krizhevsky, G.~Hinton \emph{et~al.}, ``Learning multiple layers of features from tiny images,'' 2009.

\bibitem{selvaraju2017grad}
R.~R. Selvaraju, M.~Cogswell, A.~Das, R.~Vedantam, D.~Parikh, and D.~Batra, ``Grad-cam: Visual explanations from deep networks via gradient-based localization,'' in \emph{Proceedings of the IEEE international conference on computer vision}, 2017, pp. 618--626.

\bibitem{yuan2022self}
H.~Yuan, S.~Chan, A.~P. Creagh, C.~Tong, D.~A. Clifton, and A.~Doherty, ``Self-supervised learning for human activity recognition using 700,000 person-days of wearable data,'' \emph{arXiv preprint arXiv:2206.02909}, 2022.

\bibitem{tang2021selfhar}
C.~I. Tang, I.~Perez-Pozuelo, D.~Spathis, S.~Brage, N.~Wareham, and C.~Mascolo, ``Selfhar: Improving human activity recognition through self-training with unlabeled data,'' \emph{arXiv preprint arXiv:2102.06073}, 2021.

\bibitem{wang2021sensor}
J.~Wang, T.~Zhu, J.~Gan, H.~Ning, and Y.~Wan, ``Sensor data augmentation with resampling for contrastive learning in human activity recognition,'' \emph{arXiv preprint arXiv:2109.02054}, 2021.

\bibitem{khaertdinov2021contrastive}
B.~Khaertdinov, E.~Ghaleb, and S.~Asteriadis, ``Contrastive self-supervised learning for sensor-based human activity recognition,'' in \emph{2021 IEEE International Joint Conference on Biometrics (IJCB)}.\hskip 1em plus 0.5em minus 0.4em\relax IEEE, 2021, pp. 1--8.

\bibitem{haresamudram2021contrastive}
H.~Haresamudram, I.~Essa, and T.~Pl{\"o}tz, ``Contrastive predictive coding for human activity recognition,'' \emph{Proceedings of the ACM on Interactive, Mobile, Wearable and Ubiquitous Technologies}, vol.~5, no.~2, pp. 1--26, 2021.

\bibitem{haresamudram2020masked}
H.~Haresamudram, A.~Beedu, V.~Agrawal, P.~L. Grady, I.~Essa, J.~Hoffman, and T.~Pl{\"o}tz, ``Masked reconstruction based self-supervision for human activity recognition,'' in \emph{Proceedings of the 2020 international symposium on wearable computers}, 2020, pp. 45--49.

\bibitem{xu2021limu}
H.~Xu, P.~Zhou, R.~Tan, M.~Li, and G.~Shen, ``Limu-bert: Unleashing the potential of unlabeled data for imu sensing applications,'' in \emph{Proceedings of the 19th ACM Conference on Embedded Networked Sensor Systems}, 2021, pp. 220--233.

\bibitem{grill2020bootstrap}
J.-B. Grill, F.~Strub, F.~Altch{\'e}, C.~Tallec, P.~Richemond, E.~Buchatskaya, C.~Doersch, B.~Avila~Pires, Z.~Guo, M.~Gheshlaghi~Azar \emph{et~al.}, ``Bootstrap your own latent-a new approach to self-supervised learning,'' \emph{Advances in neural information processing systems}, vol.~33, pp. 21\,271--21\,284, 2020.

\bibitem{shin2022mydj}
J.~Shin, S.~Lee, T.~Gong, H.~Yoon, H.~Roh, A.~Bianchi, and S.-J. Lee, ``Mydj: Sensing food intakes with an attachable on your eyeglass frame,'' in \emph{CHI Conference on Human Factors in Computing Systems}, 2022, pp. 1--17.

\bibitem{jain2022collossl}
Y.~Jain, C.~I. Tang, C.~Min, F.~Kawsar, and A.~Mathur, ``Collossl: Collaborative self-supervised learning for human activity recognition,'' \emph{Proceedings of the ACM on Interactive, Mobile, Wearable and Ubiquitous Technologies}, vol.~6, no.~1, pp. 1--28, 2022.

\bibitem{deldari2022cocoa}
S.~Deldari, H.~Xue, A.~Saeed, D.~V. Smith, and F.~D. Salim, ``Cocoa: Cross modality contrastive learning for sensor data,'' \emph{Proceedings of the ACM on Interactive, Mobile, Wearable and Ubiquitous Technologies}, vol.~6, no.~3, pp. 1--28, 2022.

\bibitem{radu2019vision2sensor}
V.~Radu and M.~Henne, ``Vision2sensor: Knowledge transfer across sensing modalities for human activity recognition,'' \emph{Proceedings of the ACM on Interactive, Mobile, Wearable and Ubiquitous Technologies}, vol.~3, no.~3, pp. 1--21, 2019.

\bibitem{bhalla2021imu2doppler}
S.~Bhalla, M.~Goel, and R.~Khurana, ``Imu2doppler: Cross-modal domain adaptation for doppler-based activity recognition using imu data,'' \emph{Proceedings of the ACM on Interactive, Mobile, Wearable and Ubiquitous Technologies}, vol.~5, no.~4, pp. 1--20, 2021.

\end{thebibliography}

\vspace{-30pt}

\begin{IEEEbiography}[{\includegraphics[width=1in,height=1.25in,clip,keepaspectratio]{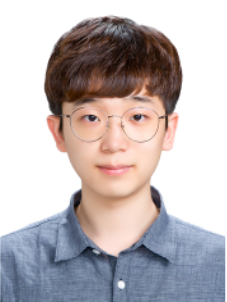}}]{Hyungjun Yoon}
received the BS (cum laude) degree in computer science from KAIST. He is currently working toward a PhD degree in electrical engineering at KAIST. His research interests include mobile computing, ubiquitous sensing, and applied machine learning.
\end{IEEEbiography}
\vspace{-30pt}

\begin{IEEEbiography}[{\includegraphics[width=1in,height=1.25in,clip,keepaspectratio]{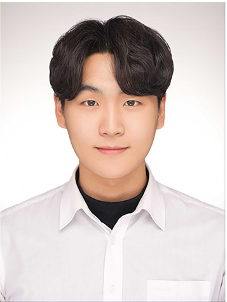}}]{Hyeongheon Cha}
received the BS (magna cum laude) degree in electrical engineering from KAIST. He is working toward a PhD in electrical engineering at KAIST. His research interests include on-device AI, mobile computing, ubiquitous sensing, and applied machine learning.
\end{IEEEbiography}
\vspace{-30pt}

\begin{IEEEbiography}[{\includegraphics[width=1in,height=1.25in,clip,keepaspectratio]{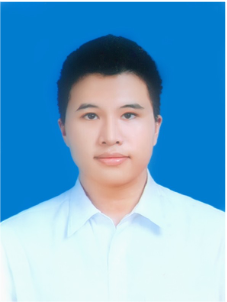}}]{Hoang C. Nguyen}
received a BS degree in computer science and mathematics from KAIST. He is a research assistant at VinUni-Illinois Smart Health Center (VISHC). His research interests include computer vision, multi-modal learning, explainable AI, and smart health.
\end{IEEEbiography}
\vspace{-30pt}

\begin{IEEEbiography}[{\includegraphics[width=1in,height=1.25in,clip,keepaspectratio]{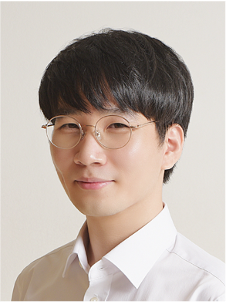}}]{Taesik Gong}
received his PhD in computer science from KAIST in 2023.  During his PhD, he interned at Google Research, Microsoft Research, and Nokia Bell Labs. He is a research scientist at Nokia Bell Labs and a visiting scholar at the University of Cambridge. His research interests include on-device AI, human-centered AI, and ubiquitous computing. He won the Best PhD Dissertation Award from the School of Computing and College of Engineering at KAIST and is also a recipient of the Google PhD Fellowship.
\end{IEEEbiography}
\vspace{-30pt}

\begin{IEEEbiography}[{\includegraphics[width=1in,height=1.25in,clip,keepaspectratio]{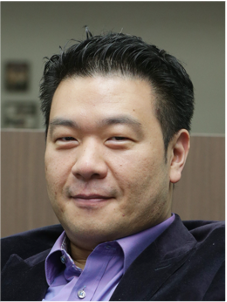}}]{Sung-Ju Lee}
received a PhD degree in computer science from the University of California, Los Angeles, California, in 2000. After spending 15 years in the industry in Silicon Valley, he joined KAIST, where he is a professor and KAIST endowed chair professor. His research interests include mobile computing, mobile ML/AI, wireless networks, and HCI. He won the HP CEO Innovation Award, the Best Paper Awards at ACM CSCW 2021 and IEEE ICDCS 2016, and the Test-of-Time Paper Award at ACM WINTECH 2016. 
\end{IEEEbiography}

\vfill

\end{document}